\documentclass[review,11pt,authoryear]{elsarticle}
\usepackage[margin=2.5cm]{geometry}

\usepackage{lineno}
\usepackage{graphicx}
\usepackage{booktabs}
\usepackage{amsmath}
\usepackage{amssymb}
\usepackage{hyperref}
\usepackage{multirow}
\usepackage{xcolor}
\usepackage{subcaption}
\usepackage{algorithm}
\usepackage{algorithmic}
\usepackage{placeins}

\modulolinenumbers[1]

\journal{International Journal of Applied Earth Observation and Geoinformation}

\begin{document}

\begin{frontmatter}

\title{Remote SAMsing: From Segment Anything to Segment Everything}

\author[ee]{Osmar Luiz Ferreira de Carvalho}
\ead{osmarcarvalho@ieee.org}

\author[geo]{Osmar Ab\'{i}lio de Carvalho J\'{u}nior\corref{cor1}}
\ead{osmarjr@unb.br}

\author[geo]{Anesmar Olino de Albuquerque}
\ead{anesmar@ieee.org}

\author[ee]{Daniel Guerreiro e Silva}
\ead{danielgs@unb.br}

\cortext[cor1]{Corresponding author}
\address[ee]{Department of Electrical Engineering, University of Bras\'{i}lia, Bras\'{i}lia, Brazil}
\address[geo]{Department of Geography, University of Bras\'{i}lia, Bras\'{i}lia, Brazil}

\begin{abstract}
SAM2 produces high-quality zero-shot segmentation on natural images, but applying it to large remote sensing scenes exposes two problems: (1) its mask generator faces an inherent quality-coverage trade-off: strict thresholds yield precise masks but leave most of the image unsegmented, while relaxed thresholds increase coverage at the cost of mask quality; and (2) large images must be tiled, fragmenting objects across tile boundaries. We propose Remote SAMsing, an open-source pipeline that solves both problems without modifying SAM2 or requiring training data. For coverage, a multi-pass algorithm runs SAM2 repeatedly on each tile, painting accepted masks black between passes to simplify the scene for the next iteration, and relaxing quality thresholds only when coverage gains stagnate, ensuring that the most precise masks are always captured first. For spatial consistency, contextual padding and a parameter-free best-match merge reconstruct objects fragmented across tile boundaries. Evaluated on seven scenes (5~cm to 4.78~m GSD), the pipeline raises coverage from 30--68\% (single-pass SAM2) to 91--98\%. Ablation experiments quantify the contribution of each component to coverage and detection quality. Per-class evaluation shows that SAM2 transfers well to discrete RS objects (buildings 95\%, cars 82--93\% Det@0.5) with segment boundaries 3--8$\times$ more precise than SLIC and Felzenszwalb baselines. Tile size functions as an implicit scale parameter: reducing it from $1{,}000$ to 250 raises Det@0.5 from 56\% to 85\%, outperforming SAM2's built-in multi-scale mechanism. The pipeline generalizes to MNF false-color imagery without retraining (99.5\% ASA) and scales to production-sized images: a 1.94 billion pixel Potsdam mosaic achieved 97\% coverage without quality degradation.
\end{abstract}


\end{frontmatter}


\section{Introduction}
\label{sec:introduction}

Within the Object-Based Image Analysis (OBIA) paradigm \citep{blaschke2010obia, hay2008geobia, blaschke2014geobia}, image segmentation is a prerequisite for thematic mapping and spatial analysis of remote sensing (RS) data. RS imagery spans a wide range of sensors, spatial resolutions, and landscape types \citep{zhu2017deep_rs, li2024dl_seg_review}, and building specialized segmentation models for each combination of these factors is costly and requires per-scenario annotation. Foundation models \citep{bommasani2021foundation, xiao2025foundation_rs} avoid this requirement by generalizing across domains from a single pretraining. The Segment Anything Model 2 (SAM2) \citep{ravi2024sam2}, successor to SAM \citep{kirillov2023segment}, produces fully unsupervised segmentation through its Automatic Mask Generator (AMG), and its learned representations show promise for RS imagery, though with known limitations at certain scales and modalities \citep{ren2024segment_anything_everywhere, osco2023sam_rs_review}.

Despite this promise, most existing work concentrates on improving per-object segmentation quality through fine-tuning, prompt engineering, or architectural adaptation on individual image patches. Tools that support large images process each tile independently without coverage optimization or boundary reconciliation. The end-to-end problem of segmenting a large RS image into a spatially consistent, complete output remains open.

This gap persists due to two problems inherent in applying SAM2 to large RS images. First, SAM2's AMG filters mask predictions by predicted IoU and stability score, and a single pass either leaves most of the image unmasked under strict thresholds or accepts low-quality masks under relaxed thresholds. Second, aerial and satellite sensors produce images spanning thousands to tens of thousands of pixels per dimension, while SAM2 internally resizes all inputs to $1{,}024 \times 1{,}024$. Dividing the image into tiles introduces boundary artifacts: objects that cross tile edges produce fragmented masks that must be reconciled into a globally consistent segmentation.

This paper presents Remote SAMsing, an open-source pipeline that addresses both problems without modifying SAM2's architecture or training task-specific prompt generators:

\begin{enumerate}
    \item \textbf{Coverage.} A multi-pass algorithm with black mask and adaptive threshold decay breaks the coverage-quality trade-off: strict thresholds capture the most salient objects first, and relaxation occurs only when progress stagnates, so the majority of segments retain high quality while achieving near-complete coverage.
    \item \textbf{Spatial consistency.} Contextual padding ensures accurate segmentation near tile edges, and a parameter-free best-match merge with Union-Find unifies boundary fragments in linear time, producing a single consistent label map for arbitrarily large images with constant GPU memory.
    \item \textbf{Experimental analysis.} No existing work evaluates how SAM2's configuration affects segmentation quality on large RS images. A systematic analysis across seven scenes spanning three spatial resolutions (5~cm to 4.78~m), two spectral compositions, and two landscape types characterizes the effects of tile size, threshold configuration, and spectral composition on per-class segmentation quality.
\end{enumerate}

The remainder of this paper reviews related work (Section~\ref{sec:related}), describes the methodology (Section~\ref{sec:methodology}), presents the experimental setup (Section~\ref{sec:experiments}) and results (Section~\ref{sec:results}), discusses their implications (Section~\ref{sec:discussion}), and concludes (Section~\ref{sec:conclusion}).

\section{Related Work}
\label{sec:related}

\subsection{SAM in Remote Sensing}

SAM \citep{kirillov2023segment} performs prompt-based image segmentation using a Vision Transformer \citep{dosovitskiy2021vit}, with the Automatic Mask Generator (AMG) placing a regular grid of points and filtering masks by predicted IoU and stability score, which control output density and quality. SAM2 \citep{ravi2024sam2} replaces the encoder with a Hiera backbone \citep{ryali2023hiera}, improving efficiency and mask quality. Both models resize inputs to $1{,}024 \times 1{,}024$ pixels and share the same AMG interface. Lightweight variants such as EfficientSAM \citep{xiong2024efficientsam} and FastSAM \citep{zhao2023fastsam} reduce computational cost but retain this fixed resolution, so scaling limitations persist for large RS images \citep{zhang2025sam2survey}.

Recent reviews \citep{osco2023sam_rs_review, wan2025samrs_survey} indicate that SAM applications in RS remain largely patch-based, focusing on task-specific adaptations such as prompt generation \citep{chen2024rsprompter}, dataset construction \citep{wang2024samrs}, and change detection \citep{ding2024samcd}. \citet{liu2026asamps} proposed an adaptive SAM2 pipeline for planted field segmentation, but it remains patch-based without addressing coverage or cross-tile consistency. Similarly, tools such as Geo-SAM \citep{zhao2023geosam} and segment-geospatial \citep{wu2023samgeo} either require manual prompting or process tiles independently, without boundary reconciliation or coverage optimization. Zero-shot evaluations further highlight these limitations, showing degraded performance on small or visually simple objects and on classes without clear instance boundaries \citep{ren2024segment_anything_everywhere, osco2023sam_rs_review}.

SAM's single-pass AMG leaves large portions of RS images unsegmented, particularly for small, complex, or sparsely sampled objects. Although a built-in multi-scale mechanism (crop\_n\_layers) exists, it operates without iterative refinement or coverage feedback, and its effectiveness for RS remains unclear. Existing methods primarily improve segment quality rather than coverage \citep{fan2025stablesam, lin2025samrefiner}. \citet{osco2023sam_rs_review} described an iterative object extraction loop using text prompts, and \citet{shepherd2019large_scale} proposed iterative elimination for traditional RS segmentation, but neither applies a multi-pass strategy to SAM's AMG for exhaustive coverage. To the best of the authors' knowledge, no prior work has addressed coverage maximization through a multi-pass strategy in SAM-based segmentation.

\subsection{Tiling and Boundary Reconciliation}

Tiling is commonly used to process large RS images, but objects crossing tile boundaries produce fragmented masks that must be reconciled into a consistent segmentation. Existing stitching and mosaicking approaches \citep{huang2018tiling, carvalho2021instance} address boundary artifacts in CNN-based predictions but are not designed for exhaustive pixel-level segmentation. Traditional region-merging methods \citep{lassalle2015scalable, lv2025deepmerge} rely on spectral similarity and do not extend naturally to the mask-based outputs of foundation models such as SAM. As a result, no existing approach provides a solution for generating spatially consistent, large-scale segmentations compatible with SAM.

If high coverage is achieved, SAM-based segments can function as superpixels for OBIA \citep{blaschke2010obia, benz2004ecognition, baatz2000multiresolution}. Traditional superpixel methods \citep{ren2003superpixel, achanta2012slic, felzenszwalb2004efficient} produce exhaustive and relatively uniform partitions, whereas SAM-based segmentation does not guarantee complete coverage and yields highly heterogeneous regions. Recent approaches, such as Superpixel Anything \citep{walther2025superpixelanything}, highlight the potential of foundation models, but still assume complete segmentation. Consequently, existing evaluation protocols \citep{stutz2018superpixels} do not assess whether full image coverage is achieved.

\section{Methodology}
\label{sec:methodology}

Remote SAMsing divides a large RS image into tiles, segments each tile through multiple SAM2 passes that progressively simplify the scene by masking already-segmented regions and relaxing thresholds only when coverage stagnates (Section~\ref{sec:method_multipass}), and merges the results across tile boundaries through contextual padding and a parameter-free best-match strategy (Section~\ref{sec:method_scaling}).

\subsection{Multi-Pass Adaptive Segmentation}
\label{sec:method_multipass}

The first pass runs SAM2's AMG on the unmodified tile with a uniform $k \times k$ point grid at the strictest thresholds ($\tau = \tau_{\text{start}}$). Subsequent passes simplify the scene and re-run SAM2 on residual areas (Section~\ref{sec:method_pass}), with thresholds relaxing only when progress stagnates (Section~\ref{sec:method_threshold}). The loop continues until coverage reaches the target, thresholds are exhausted, or the maximum number of passes is reached. Algorithm~\ref{alg:multipass} formalizes the procedure.

\begin{algorithm}[h!]
\caption{Multi-pass adaptive segmentation (per tile)}
\label{alg:multipass}
\begin{algorithmic}[1]
\REQUIRE Tile image $I$, thresholds $\tau_{\text{start}}$ and $\tau_{\text{end}}$, decay step $\Delta$, stagnation threshold $\varepsilon$, target coverage
\ENSURE Label map $L$
\STATE $\tau \leftarrow \tau_{\text{start}}$
\STATE Run SAM2 on $I$ with $k \times k$ point grid at $\tau$; assign labels
\REPEAT
    \STATE Paint segmented pixels black in $I$
    \STATE Place prompts on residual areas
    \STATE Run SAM2 at $\tau$; accept and assign new masks
    \IF{coverage gain $< \varepsilon$}
        \STATE $\tau \leftarrow \tau - \Delta$
    \ENDIF
\UNTIL{coverage $\geq$ target \OR $\tau < \tau_{\text{end}}$}
\RETURN $L$
\end{algorithmic}
\end{algorithm}

\subsubsection{Per-Pass Processing}
\label{sec:method_pass}
\label{sec:method_blackmask}

After each pass, all segmented pixels are painted black, erasing captured objects from the image. The scene grows simpler with each iteration: objects that were previously too subtle for SAM2 to detect become the most prominent features in what remains. This simplification operates through two mechanisms: SAM2 generates few or no mask candidates over uniform black regions, preventing repeated segmentation of the same objects, and the abrupt transition between the black mask and the remaining content creates an artificial boundary that SAM2 treats as an object edge, so new masks naturally complement rather than overlap with existing ones.

Prompts are then placed only on residual areas. The first pass uses a uniform $k \times k$ grid. Subsequent passes regenerate the same grid but discard all points that fall on already-segmented pixels. This Dense Grid strategy maintains uniform point density across all residual regions, ensuring that small isolated fragments receive prompts even as the residual area shrinks across passes.

SAM2 masks may contain disconnected regions, so each mask is split into its connected components and evaluated independently. Components smaller than $a_{\min}$ pixels are discarded, and components where more than half of the pixels overlap with existing segments are rejected as redundant. This prevents a useful mask from being discarded just because one of its disconnected fragments overlaps with an existing segment.

\subsubsection{Adaptive Threshold Decay}
\label{sec:method_threshold}

SAM2's AMG filters masks by two quality scores: predicted IoU ($\tau_{\text{iou}}$) and stability ($\tau_{\text{stab}}$). The initial pass uses high thresholds ($\tau^{\text{start}}$), corresponding to the upper range of SAM2's internal quality scores where virtually all accepted masks are spatially coherent. When a pass does not produce new masks or coverage gain falls below a stagnation threshold, both scores are reduced by a fixed step $\Delta$. This continues until the thresholds reach $\tau^{\text{end}}$, below which SAM2 produces predominantly noisy masks. Threshold relaxation occurs only when progress stagnates, preserving mask quality in earlier passes.

\subsection{Scaling to Large Images}
\label{sec:method_scaling}

\subsubsection{Tiling and Contextual Padding}
\label{sec:method_tiling}

SAM2 occasionally stops masks a few pixels before the image edge, leaving a narrow unmasked strip that prevents contact between segments in adjacent tiles. Contextual padding solves this: the input image is divided into non-overlapping tiles of $T \times T$ pixels, but each tile is extracted with $p$ additional pixels on each side, producing an inference window of $(T + 2p)^2$ pixels. After segmentation, the padded margin is discarded and only the central $T \times T$ core is kept. Segments that extended into the padded region are cut at the tile boundary, guaranteeing contact between adjacent tiles and enabling the merge algorithm described below. The default values of $T$ and $p$ are listed in Table~\ref{tab:default_config}.

\subsubsection{Boundary Merge}
\label{sec:method_merge}

Despite padding, tiles are segmented independently, so the same object receives different labels on each side of a tile boundary. The challenge is deciding which pairs to merge: merging all touching pairs propagates spurious fusions (e.g., a building and an adjacent road sharing a few pixels of contact would be joined into a single segment). Remote SAMsing employs a parameter-free \textit{best-match} strategy: for each tile boundary, the algorithm records the contact count between every pair of adjacent labels, and each segment selects the neighbor with the highest contact. Only these best-match pairs are merged. This handles both 1:1 splits (one object across two tiles) and $N$:1 splits (one object fragmented on one side but whole on the other), while each segment commits to exactly one merge partner. A stricter mutual-best criterion would fail on asymmetric splits, and a contact-threshold criterion would require a tunable parameter.

The merge pairs are processed by a Union-Find structure \citep{tarjan1975union} with path compression, which handles transitive merges implicitly and runs in linear time. After merging, small enclosed components are absorbed into their surrounding segment and isolated fragments below $a_{\min}$ pixels (Table~\ref{tab:default_config}) are removed as noise.

\section{Experimental Setup}
\label{sec:experiments}

The evaluation assesses the pipeline contributions (coverage and spatial consistency) and characterizes how SAM2's configuration affects segmentation quality on large RS images.

\subsection{Study Areas and Datasets}
\label{sec:exp_data}

The evaluation uses seven scenes from three datasets spanning different sensors, spatial resolutions, spectral compositions, and land cover types (Table~\ref{tab:datasets}).

\begin{table}[htbp]
\centering
\caption{Overview of datasets used in the experiments. Tile counts refer to the default tile size of $1000 \times 1000$~pixels.}
\label{tab:datasets}
\begin{tabular*}{\textwidth}{@{\extracolsep{\fill}}llllrrl}
\toprule
Dataset & Sensor & Composition & Res.\ (m) & Scenes & Size (px) & Tiles \\
\midrule
ISPRS Potsdam & Aerial & RGB natural & 0.05 & 3 & 6{,}000$^2$ & 36 \\
Bras\'{i}lia & Aerial & RGB natural & 0.24 & 3 & 8{,}000$^2$ & 64 \\
Agri-BR & Planet & MNF false-color & 4.78 & 1 & 10{,}000$^2$ & 100 \\
\bottomrule
\end{tabular*}
\end{table}

\textbf{ISPRS Potsdam.} Three patches (3\_13, 5\_12, 5\_13) from the ISPRS 2D Semantic Labeling Contest \citep{rottensteiner2014isprs}, covering a dense European urban landscape at 5~cm GSD. At this resolution, objects are detailed but amorphous land cover classes (low vegetation, impervious surfaces) challenge boundary delineation. Public semantic ground truth includes six classes: impervious surfaces, buildings, low vegetation, trees, cars, and clutter.

\textbf{Bras\'{i}lia.} Three crops from a high-resolution aerial survey of Bras\'{i}lia, Brazil \citep{carvalho2022panoptic}, representing distinct urban morphologies at 24~cm GSD: residential (BSB-1), commercial (BSB-2), and mixed (BSB-3). At this resolution, individual cars and sidewalks are visible, producing a high density of small objects that test per-object detection. BSB-1 includes instance-level ground truth with nine classes: buildings, trees, cars, pools, courts, decks, roads, lakes, and permeable surfaces.

\textbf{Agri-BR.} A crop from a Planet satellite image \citep{frazier2021planet} over an agricultural region in central Brazil, using a false-color composition from MNF transformation \citep{green1988mnf} at 4.78~m GSD. This dataset tests generalization to non-RGB imagery and large homogeneous objects. Ground truth includes three classes: pivot irrigation, crop fields, and lakes.

\subsection{Implementation Details}
\label{sec:exp_impl}

All experiments use the SAM2.1 model with the Hiera Large backbone, the largest available checkpoint. The pipeline is implemented in Python~3.12 using PyTorch~2.9 with CUDA~12.8 for GPU inference. All experiments were run on a workstation equipped with an Intel Core i9-14900K CPU, 64~GB of RAM, and an NVIDIA RTX~4090 GPU with 24~GB of VRAM. The default configuration parameters are listed in Table~\ref{tab:default_config}. The source code is publicly available at \url{https://github.com/osmarluiz/sam-mosaic}.

\begin{table}[htbp]
\centering
\caption{Default pipeline configuration parameters.}
\label{tab:default_config}
\scriptsize
\begin{tabular*}{\textwidth}{@{\extracolsep{\fill}}llrl}
\toprule
Module & Parameter & Default & Description \\
\midrule
Tiling & $T$ (tile\_size) & 1{,}000 & Tile side length (pixels) \\
       & $p$ (padding) & 50 & Contextual padding (pixels) \\
\midrule
Segmentation & $k$ (points\_per\_side) & 64 & Grid density ($k^2$ points) \\
             & $\text{cov}_{\text{target}}$ & 99\% & Target coverage \\
             & $\tau_{\text{iou}}^{\text{start}}$ / $\tau_{\text{iou}}^{\text{end}}$ & 0.93 / 0.60 & IoU threshold range \\
             & $\tau_{\text{stab}}^{\text{start}}$ / $\tau_{\text{stab}}^{\text{end}}$ & 0.93 / 0.60 & Stability threshold range \\
             & $\Delta$ (step) & 0.01 & Threshold decay step \\
             & $\varepsilon$ (stagnation) & 0.1 pp & Min.\ coverage gain to keep $\tau$ \\
             & Overlap rejection & 50\% & Max.\ overlap with existing segments \\
\midrule
Merge & Strategy & best\_match & Parameter-free best-match \\
\midrule
Post-proc. & $a_{\min}$ (min\_mask\_area) & 100 & Min.\ segment area (pixels) \\
           & merge\_enclosed\_max & 500 & Max.\ area for enclosed absorption \\
\bottomrule
\end{tabular*}
\end{table}

\subsection{Experimental Design}
\label{sec:exp_design}

\label{sec:exp_metrics} All experiments are evaluated using the metrics in Table~\ref{tab:metrics}. Since SAM2 segments visually coherent regions rather than semantic objects, a single building may produce separate roof and shadow segments without this being a failure. Per-object metrics therefore use a greedy oracle protocol adapted from \citet{clinton2010segmentation_eval}: for each GT polygon, segments are greedily combined to maximize the reconstructed IoU, so an over-segmented object achieves high IoU as long as its fragments together cover the GT polygon without leaking into adjacent classes. All experiments are fully deterministic, so variance estimates are unnecessary.

\begin{table}[htbp]
\centering
\caption{Evaluation metrics.}
\label{tab:metrics}
\scriptsize
\begin{tabular*}{\textwidth}{@{\extracolsep{\fill}}lp{11cm}}
\toprule
Metric & Description \\
\midrule
Coverage (\%) & Fraction of image pixels assigned to any segment \\
ASA \citep{stutz2018superpixels} & Fraction of pixels correctly labeled when each segment receives its majority GT class \\
Det@0.5 (\%) & Fraction of GT objects reconstructed with greedy oracle IoU $\geq 0.5$ \\
SS-Det@0.5 (\%) & Fraction of GT objects matched by a single segment with IoU $\geq 0.5$ \\
mIoU & Mean greedy oracle IoU across all GT objects \\
$\bar{n}$ & Mean segments per GT object (over-segmentation indicator) \\
BIoU \citep{cheng2021boundary} & IoU restricted to a $d = 3$ pixel band around boundaries \\
\bottomrule
\end{tabular*}
\end{table}

The following ablation studies isolate the contribution of each pipeline component:

\begin{itemize}
    \item \textit{Multi-pass contribution:} single-pass SAM2 at $\tau = 0.93$, $0.88$, $0.70$ vs.\ full pipeline.
    \item \textit{Native multi-scale:} single-pass with crop\_n\_layers $= 1$ at $\tau = 0.88$ and $0.70$.
    \item \textit{Component ablation:} full pipeline with black mask off; adaptive decay off.
    \item \textit{Tile size effect:} $T \in \{1{,}000, 500, 250\}$ ($1\times$, $2\times$, $4\times$ magnification).
    \item \textit{Existing SAM tools:} SamGeo2 \citep{wu2023samgeo} with default configuration.
    \item \textit{Merge strategy:} best-match vs.\ naive across all seven scenes.
\end{itemize}

Beyond coverage, the evaluation asks whether the resulting segments are useful: do they align with real objects, respect class boundaries, and outperform traditional alternatives? At the best tile size per scene, segmentation quality is assessed per class using the greedy oracle protocol and ASA. Remote SAMsing is then compared against SLIC \citep{achanta2012slic} and Felzenszwalb \citep{felzenszwalb2004efficient}, calibrated to approximate Remote SAMsing's segment count per dataset. Finally, a scalability test applies the full pipeline to a $36{,}000 \times 54{,}000$ pixel Potsdam mosaic (1.94 billion pixels) to verify that quality does not degrade with image size.

\section{Results}
\label{sec:results}

\begin{figure}[p]
    \centering
    \includegraphics[width=0.97\textwidth]{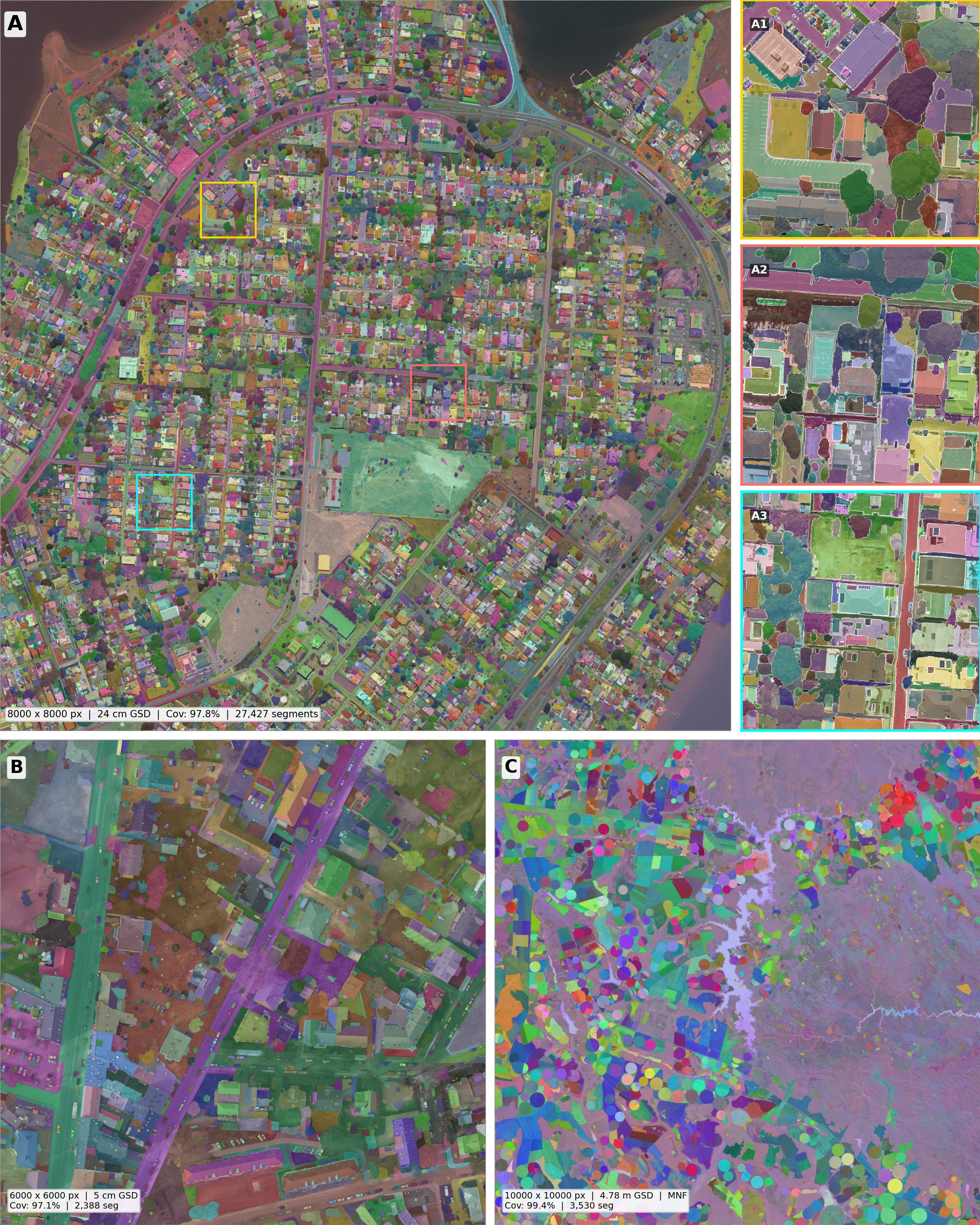}
    \caption{Full segmentation results across three datasets. (A)~BSB-1: high-resolution urban scene at 24~cm GSD (Dense Grid, $T = 250$, 27{,}427 segments, 97.8\% coverage). Panels A1--A3 show zoom details with segment boundaries. (B)~Potsdam-1: ultra-high-resolution urban scene at 5~cm GSD (Dense Grid, $T = 1{,}000$, 2{,}388 segments, 97.1\% coverage). (C)~Agri-BR: agricultural scene at 4.78~m GSD with MNF false-color composition (Dense Grid, $T = 1{,}000$, 4{,}533 segments, 98.1\% coverage). Segment colors are assigned randomly for visualization.}
    \label{fig:full_segmentation}
\end{figure}

Fig.~\ref{fig:full_segmentation} illustrates the complete pipeline output on three scenes, achieving 97--98\% coverage with segments that follow individual objects rather than arbitrary pixel boundaries. The remainder of this section examines how each pipeline component contributes to this result (Section~\ref{sec:res_config}), whether the boundary merge preserves object identity across tiles (Section~\ref{sec:res_merge}), how segment quality varies per class (Section~\ref{sec:res_quality}), how it compares to traditional methods (Section~\ref{sec:res_baselines}), and whether the pipeline scales to production-sized images (Section~\ref{sec:res_scalability}).

\subsection{Pipeline Configuration Analysis}
\label{sec:res_config}
\label{sec:res_ablation}

\begin{figure}[htbp]
    \centering
    \includegraphics[width=\textwidth]{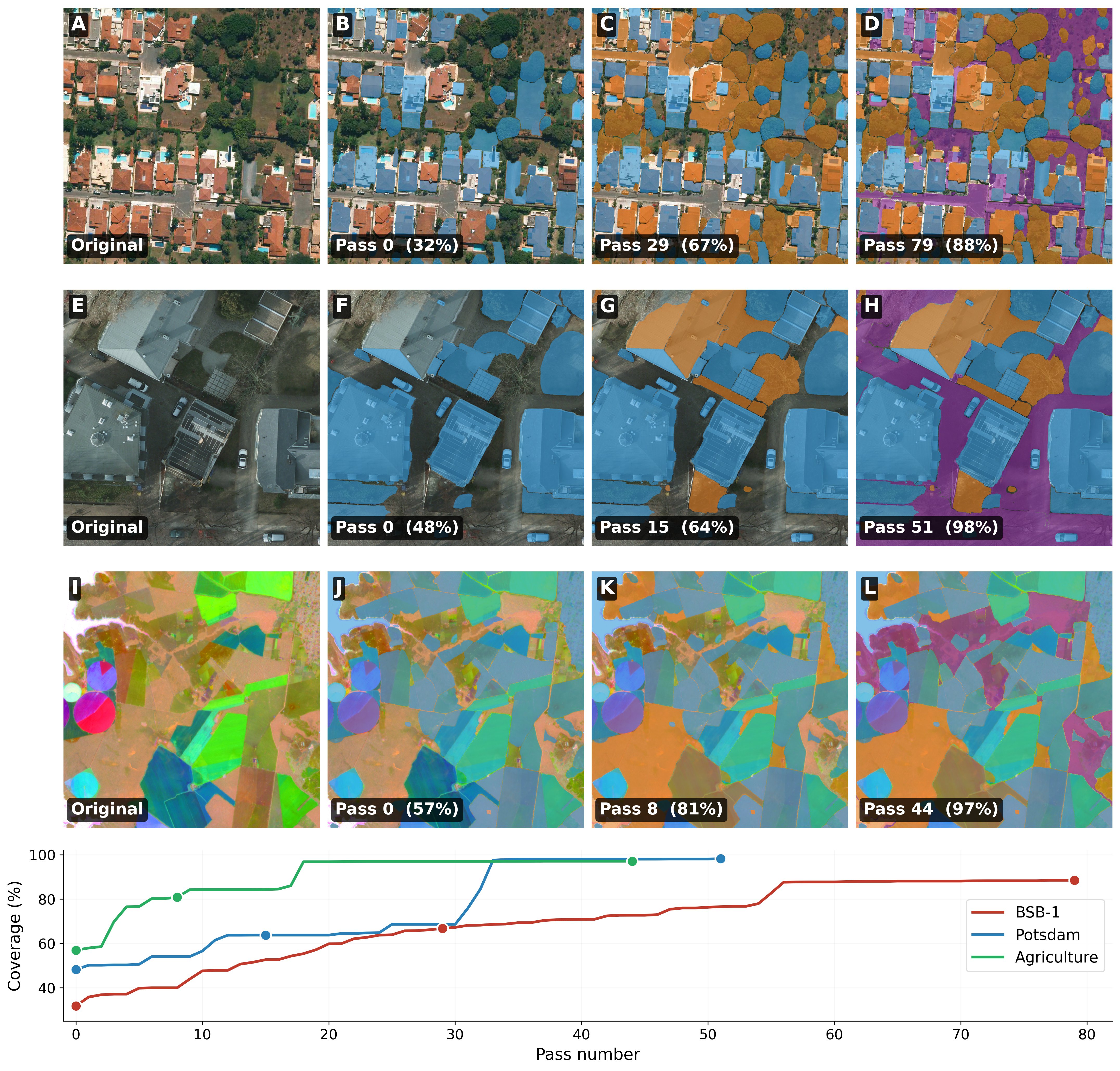}
    \caption{Multi-pass segmentation progression across three datasets. Each row shows a different scene; each column shows the cumulative result at a different stage. Segments are colored by the stage in which they were added: blue = first pass, orange = intermediate passes, purple = final passes. The bottom panel shows coverage (\%) versus pass number.}
    \label{fig:multipass_progression}
\end{figure}

Single-pass SAM2 at the highest quality threshold ($\tau = 0.93$) covers only 30--68\% of the image, with 15--76\% Det@0.5 (Table~\ref{tab:config}). Even at the most permissive threshold ($\tau = 0.70$), coverage reaches only 77--93\%. The full pipeline raises coverage to 91--98\% across all three scenes. Running the same configuration on the remaining scenes confirms this pattern within each dataset: BSB-2 and BSB-3 reach 94\% and 93\% coverage, while Potsdam-2 and Potsdam-3 reach 98\% each. For comparison, SamGeo2 \citep{wu2023samgeo}, which applies SAM2 with default tiling but no multi-pass pipeline, achieves only 6--27\% coverage. Fig.~\ref{fig:multipass_progression} illustrates the progression on individual tiles.

\begin{table}[htbp]
\centering
\caption{Pipeline configuration analysis: coverage (\%), processing time (seconds), detection rate at IoU $\geq 0.5$ (Det@0.5, \%), and Boundary IoU (BIoU) across three scenes. All configurations use Dense Grid point sampling.}
\label{tab:config}
\scriptsize
\begin{tabular*}{\textwidth}{@{\extracolsep{\fill}}l@{\hspace{6pt}}rrrr@{\hspace{8pt}}rrrr@{\hspace{8pt}}rrrr}
\toprule
 & \multicolumn{4}{c}{BSB-1 (24~cm)} & \multicolumn{4}{c}{Potsdam-1 (5~cm)} & \multicolumn{4}{c}{Agri-BR (4.78~m)} \\
\cmidrule(lr){2-5} \cmidrule(lr){6-9} \cmidrule(lr){10-13}
Configuration & Cov. & Time & Det & BIoU & Cov. & Time & Det & BIoU & Cov. & Time & Det & BIoU \\
\midrule
\multicolumn{13}{c}{\footnotesize\textit{Single pass}} \\
\midrule
SamGeo2 (default config) & 5.8 & 261 & 6.7 & .048 & 9.7 & 113 & 20.1 & .083 & 26.5 & 319 & 29.2 & .206 \\
$\tau = 0.93$ & 30.1 & 416 & 15.3 & .065 & 36.5 & 190 & 35.9 & .125 & 68.5 & 671 & 76.3 & .757 \\
$\tau = 0.88$ (SAM2 default) & 53.1 & 526 & 38.8 & .159 & 60.8 & 228 & 53.9 & .168 & 81.0 & 619 & 86.7 & .788 \\
$\tau = 0.70$ & 77.0 & 945 & 61.8 & .244 & 86.1 & 454 & 65.2 & .184 & 93.4 & 1{,}162 & 90.1 & .776 \\
\midrule
\multicolumn{13}{c}{\footnotesize\textit{Single pass + crop\_n\_layers = 1}} \\
\midrule
$\tau = 0.88$ + cnl & 56.5 & 2{,}740 & 45.5 & .205 & 61.8 & 722 & 56.2 & .181 & 82.2 & 2{,}074 & 89.2 & .586 \\
$\tau = 0.70$ + cnl & 79.0 & 3{,}483 & 72.4 & .322 & 85.1 & 1{,}336 & 70.9 & .198 & 92.6 & 4{,}421 & 94.4 & .578 \\
\midrule
\multicolumn{13}{c}{\footnotesize\textit{Multi-pass ablation ($T = 1{,}000$)}} \\
\midrule
All components & 91.5 & 13{,}065 & 56.0 & .199 & 97.1 & 2{,}549 & 61.9 & .175 & 98.1 & 3{,}959 & 95.8 & .894 \\
-- black mask & 80.4 & 11{,}165 & 65.1 & .256 & 87.7 & 3{,}575 & 69.2 & .190 & 94.4 & 4{,}986 & 94.5 & .841 \\
-- adaptive thr. & 71.8 & 1{,}611 & 51.8 & .201 & 87.6 & 478 & 64.4 & .185 & 95.5 & 1{,}093 & 95.2 & .828 \\
\midrule
\multicolumn{13}{c}{\footnotesize\textit{Tile size (all components)}} \\
\midrule
$T = 500$ & 96.9 & 24{,}872 & 73.2 & .329 & 98.8 & 5{,}896 & 55.9 & .172 & 99.1 & 6{,}988 & 82.5 & .470 \\
$T = 250$ & 97.8 & 66{,}253 & 85.1 & .543 & 99.2 & 11{,}851 & 49.5 & .164 & 99.4 & 22{,}252 & 74.6 & .407 \\
\bottomrule
\end{tabular*}
\end{table}

\begin{figure}[htbp]
    \centering
    \includegraphics[width=\textwidth]{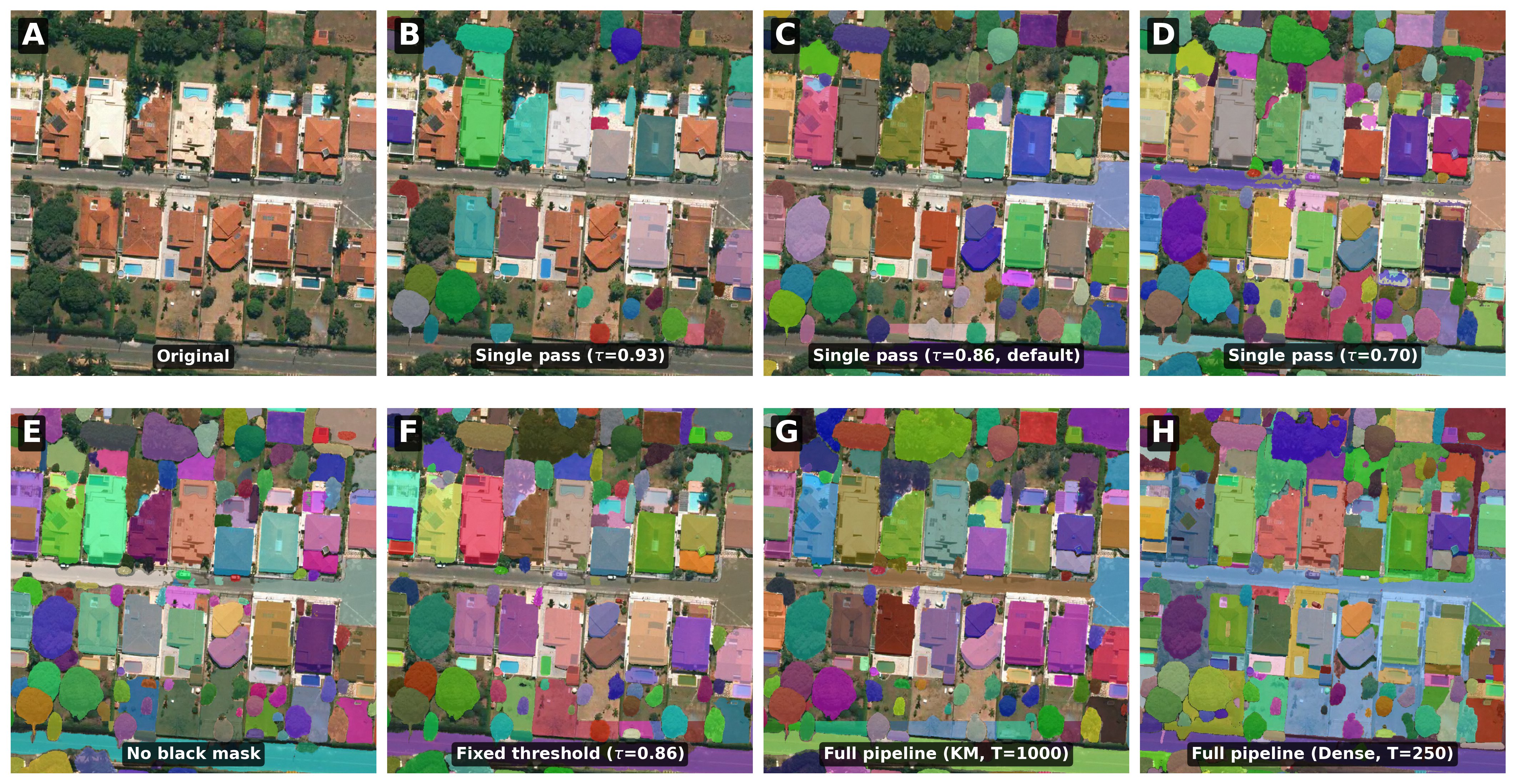}
    \caption{Visual comparison of pipeline configurations on a BSB-1 crop: (A) original image, (B) SamGeo2, (C) single-pass SAM2 at $\tau = 0.88$, (D) single-pass with crop\_n\_layers $= 1$ and $\tau = 0.70$, (E) multi-pass without black mask, (F) multi-pass with fixed threshold (no adaptive decay), (G) full pipeline at $T = 1{,}000$, and (H) full pipeline at $T = 250$.}
    \label{fig:ablation_visual}
\end{figure}

Adaptive threshold decay is the largest single contributor: disabling it reduces coverage by 3--20~pp (19.7~pp on BSB-1). The black mask adds 4--11~pp by preventing SAM2 from repeatedly generating masks for already-segmented regions. High coverage comes at a cost, however: at $T = 1{,}000$, the full pipeline covers more than the no-black-mask variant (91\% vs.\ 80\% on BSB-1) but detects fewer individual objects (56\% vs.\ 65\% Det@0.5), because late passes at relaxed thresholds produce large segments that span multiple objects (Fig.~\ref{fig:ablation_visual}).

\begin{figure}[htbp]
    \centering
    \includegraphics[width=0.97\textwidth]{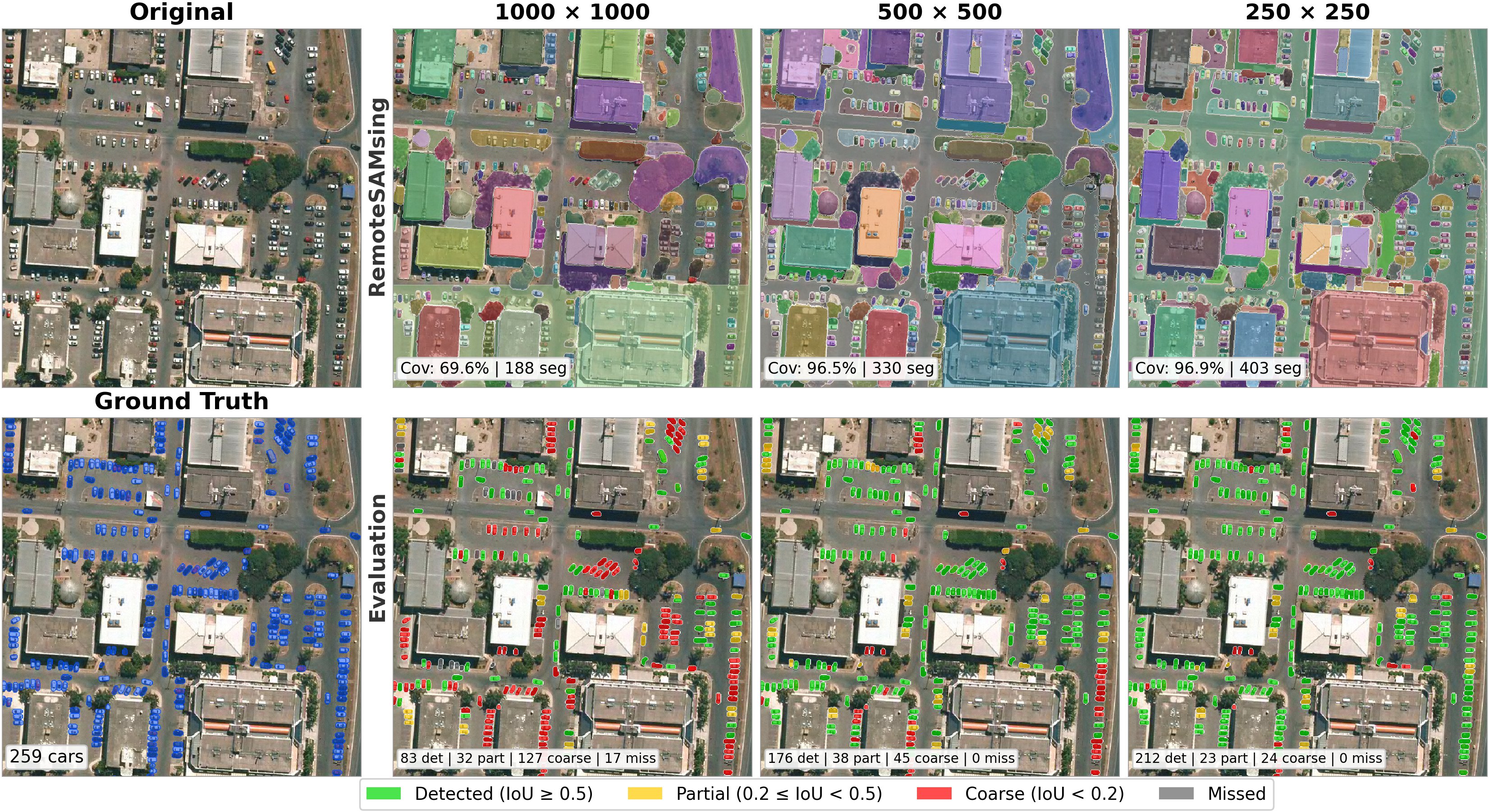}
    \caption{Effect of tile size on car detection. Top row: original image and Remote SAMsing segments at three tile sizes. Bottom row: ground truth car polygons and per-car greedy oracle evaluation. Green: detected (IoU $\geq 0.5$). Yellow: partial segmentation (0.2 $\leq$ IoU $< 0.5$). Red: coarse segmentation (IoU $< 0.2$). Gray: missed.}
    \label{fig:tilesize_cars}
\end{figure}

Reducing tile size resolves this trade-off. Smaller tiles magnify objects in SAM2's fixed $1{,}024 \times 1{,}024$ internal resolution, and on BSB-1, Det@0.5 rises from 56\% ($T = 1{,}000$) to 85\% ($T = 250$) with BIoU improving from 0.20 to 0.54. The per-car analysis (Fig.~\ref{fig:tilesize_cars}) illustrates this concretely: at $T = 1{,}000$, 51\% of cars are partially segmented, while at $T = 250$, 81\% are individually detected. SAM2's built-in multi-scale mechanism (crop\_n\_layers $= 1$) offers an alternative, but tile size reduction proves more effective: crop\_n\_layers improves Det@0.5 by 6~pp on BSB-1 at $T = 1{,}000$, while reducing tile size to $T = 250$ improves it by 29~pp at comparable processing time (18~h vs.\ 11.6~h). On Agri-BR, crop\_n\_layers also fragments large agricultural fields, decreasing BIoU from 0.78 to 0.58.

Smaller tiles do not improve quality on all scenes. On Agri-BR, BIoU drops from 0.89 ($T = 1{,}000$) to 0.40 ($T = 250$), and on Potsdam-1 Det@0.5 drops from 62\% to 50\%, as tiles covering too little ground area produce single-segment outputs that chain through the merge algorithm into mega-segments. This degradation does not occur on BSB-1, where $T = 250$ covers $60 \times 60$~m with sufficient object diversity per tile. The optimal tile size is therefore scene-dependent: the per-class evaluation in Section~\ref{sec:res_quality} adopts $T = 250$ for BSB-1 (maximizing small-object detection at 24~cm GSD) and $T = 1{,}000$ for Potsdam-1 and Agri-BR (avoiding merge chaining).

\FloatBarrier
\subsection{Boundary Merge}
\label{sec:res_merge}

\begin{figure}[htbp]
    \centering
    \includegraphics[width=0.97\textwidth]{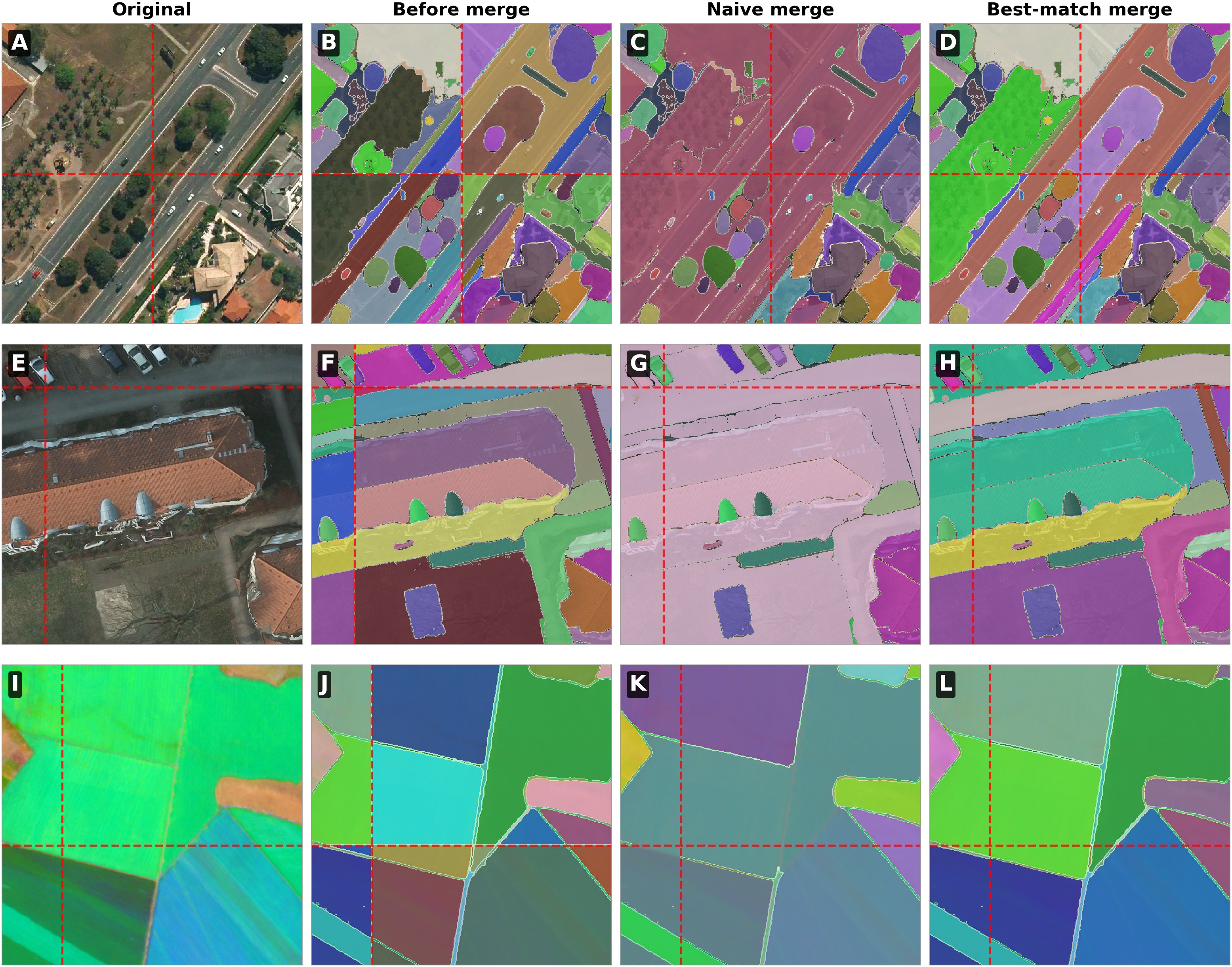}
    \caption{Boundary merge comparison across three datasets. Each row shows a crop centered on a tile boundary (red dashed lines). Before merge: independent per-tile segments with different labels on each side. Naive merge: all touching segments are merged transitively, creating spurious mega-segments. Best-match merge: each segment merges only with its highest-contact neighbor, preserving object identity.}
    \label{fig:merge}
\end{figure}

\begin{figure}[htbp]
    \centering
    \includegraphics[width=0.97\textwidth]{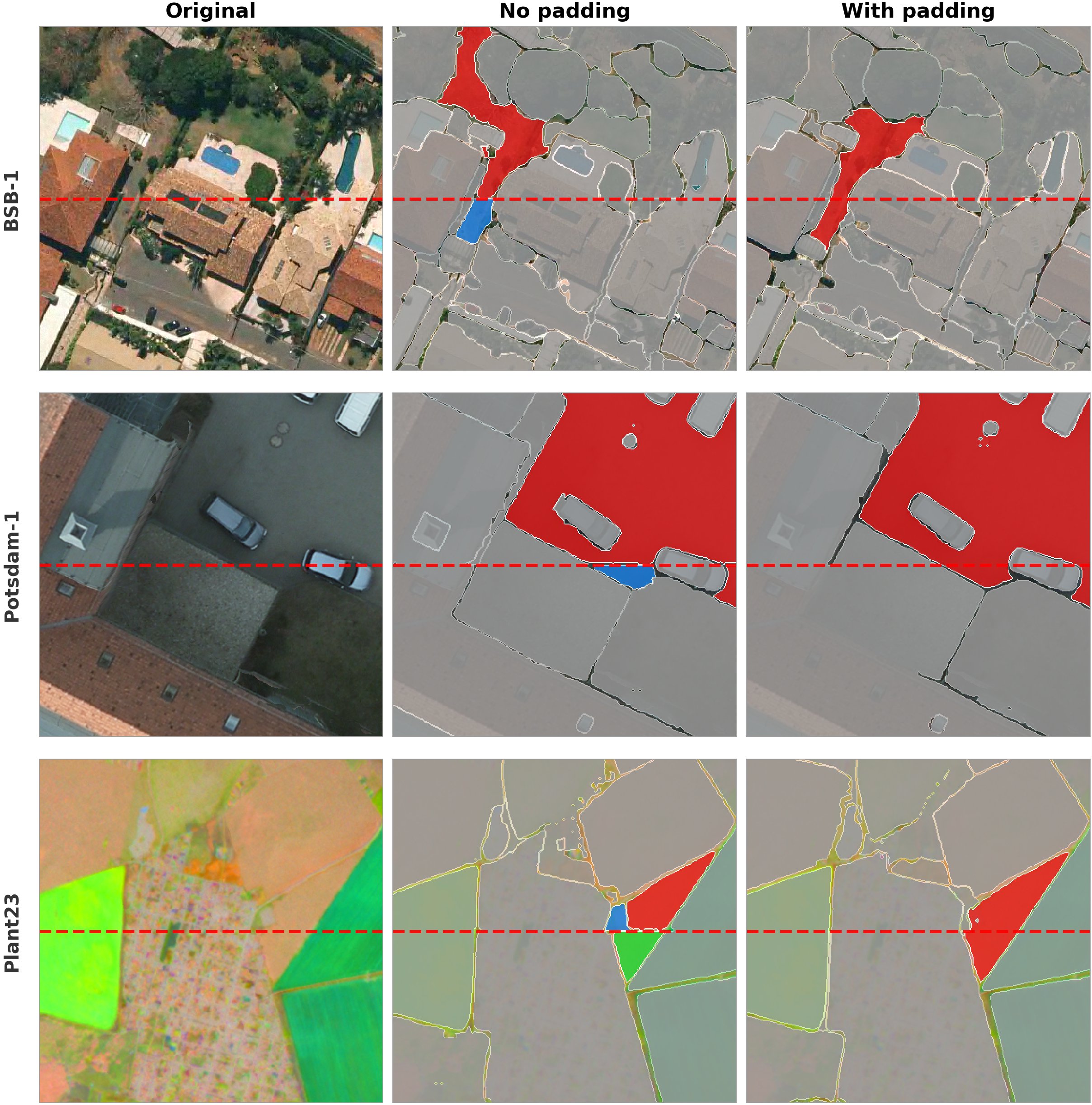}
    \caption{Effect of contextual padding on boundary segmentation. Colored segments are affected by the tile boundary (red dashed line); gray segments are unaffected. Without padding, segments terminate at the boundary, creating visible discontinuities. With padding, the same objects are segmented continuously across tile edges.}
    \label{fig:padding}
\end{figure}

The merge strategy determines whether tiled segmentation preserves object identity or collapses into unusable mega-segments (Fig.~\ref{fig:merge}). A naive baseline that merges all touching segment pairs propagates fusions transitively: on BSB-1, multiple buildings and vegetation areas collapse into a single segment spanning the tile corner, and on Potsdam-1, 83\% of the crop fuses into one mega-segment. The best-match strategy avoids this by requiring each segment to commit to exactly one merge partner, the neighbor with the highest contact area. Across all seven scenes, best-match produces 10--20\% fewer merges than naive, preserving 3--7\% more distinct segments (Table~\ref{tab:merge_strategy}). Fig.~\ref{fig:padding} shows that without padding, segments terminate before tile edges, and the merge algorithm has no contact to work with. With padding, objects are segmented continuously across tile edges, enabling correct merges.

\begin{table}[htbp]
\centering
\caption{Boundary merge comparison (Dense Grid, $T = 1{,}000$).}
\label{tab:merge_strategy}
\begin{tabular*}{\textwidth}{@{\extracolsep{\fill}}lrrrrr}
\toprule
 & \multicolumn{2}{c}{Naive} & \multicolumn{2}{c}{Best-match} & \\
\cmidrule(lr){2-3} \cmidrule(lr){4-5}
Scene & Seg. & Merges & Seg. & Merges & $\Delta$Seg \\
\midrule
Potsdam-1 & 1{,}999 & 751 & 2{,}129 & 602 & +130 \\
Potsdam-2 & 2{,}913 & 842 & 3{,}034 & 703 & +121 \\
Potsdam-3 & 2{,}389 & 668 & 2{,}482 & 554 & +93 \\
BSB-1     & 11{,}844 & 2{,}250 & 12{,}162 & 1{,}903 & +318 \\
BSB-2     & 10{,}584 & 2{,}425 & 10{,}985 & 1{,}994 & +401 \\
BSB-3     & 10{,}197 & 2{,}242 & 10{,}535 & 1{,}867 & +338 \\
Agri-BR   & 3{,}944 & 1{,}659 & 4{,}094 & 1{,}487 & +150 \\
\bottomrule
\end{tabular*}
\end{table}

\FloatBarrier
\subsection{Segmentation Quality}
\label{sec:res_quality}

Per-class evaluation at the best tile size per scene (Table~\ref{tab:perobject}) reveals that detection quality depends on object morphology: discrete objects with well-defined shapes achieve high detection (buildings 95\% Det@0.5 on both BSB-1 and Potsdam-1, cars 82--93\%, and agricultural fields 95--100\% with near-perfect one-segment-per-object correspondence ($\bar{n} \approx 1.0$)). On Agri-BR, pivot irrigation fields reach 100\% Det@0.5 with BIoU 0.98 despite the MNF false-color composition. Amorphous land cover classes score lower on Det@0.5, following the ``things'' vs.\ ``stuff'' distinction \citep{kirillov2019panoptic}: on Potsdam-1, impervious surfaces reach only 42\% and low vegetation 37\%. However, this does not indicate segmentation failure. Per-class ASA remains high (impervious 75\%, low vegetation 76\%), meaning the pixels are correctly assigned, but the elongated or irregular GT polygons are not reconstructed as single objects. Roads on BSB-1 illustrate this: 79\% Det@0.5 but 88\% ASA.

\begin{table}[htbp]
\centering
\caption{Per-class evaluation using the greedy oracle protocol. Metrics defined in Table~\ref{tab:metrics}.}
\label{tab:perobject}
\scriptsize
\begin{tabular*}{\textwidth}{@{\extracolsep{\fill}}lrrrrrrr}
\toprule
Class & $n$ & Det@0.5 & SS-Det & mIoU & BIoU & ASA & $\bar{n}$ \\
\midrule
\multicolumn{8}{c}{\footnotesize\textit{BSB-1 (Dense Grid, $T = 250$, 27{,}427 segments, 97.8\% coverage)}} \\
\midrule
\multicolumn{8}{l}{\scriptsize\textit{Things}} \\
Buildings   & 2{,}024 & 95.1 & 81.2 & 0.818 & 0.344 & 96.1 & 2.6 \\
Trees       & 8{,}767 & 87.0 & 85.8 & 0.824 & 0.618 & 92.6 & 1.1 \\
Cars        & 1{,}579 & 81.5 & 81.4 & 0.728 & 0.555 & 84.4 & 1.0 \\
Pools       & 1{,}094 & 76.8 & 75.8 & 0.700 & 0.330 & 91.1 & 1.0 \\
Courts      &      52 & 96.2 & 96.2 & 0.784 & 0.309 & 93.5 & 1.2 \\
Decks       &      10 & 80.0 & 80.0 & 0.718 & 0.434 & 88.7 & 1.0 \\
\cmidrule{1-8}
\multicolumn{8}{l}{\scriptsize\textit{Stuff}} \\
Roads       &      43 & 79.1 & 60.5 & 0.651 & 0.241 & 87.9 & 3.5 \\
Lakes       &       8 & 87.5 & 87.5 & 0.774 & 0.449 & 99.6 & 1.2 \\
Permeable   & 3{,}671 & 79.2 & 76.7 & 0.748 & 0.538 & 81.7 & 1.2 \\
\textit{Global} & \textit{17{,}248} & \textit{85.1} & \textit{82.2} & \textit{0.790} & \textit{0.543} & \textit{90.0} & --- \\
\midrule
\multicolumn{8}{c}{\footnotesize\textit{Potsdam-1 (Dense Grid, $T = 1{,}000$, 2{,}388 segments, 97.1\% coverage)}} \\
\midrule
\multicolumn{8}{l}{\scriptsize\textit{Things}} \\
Building    &      78 & 94.9 & 67.9 & 0.861 & 0.218 & 96.5 & 5.7 \\
Car         &     257 & 93.0 & 92.6 & 0.818 & 0.346 & 89.0 & 1.0 \\
Tree        &     280 & 73.2 & 59.6 & 0.582 & 0.071 & 52.2 & 2.4 \\
\cmidrule{1-8}
\multicolumn{8}{l}{\scriptsize\textit{Stuff}} \\
Impervious  &     102 & 42.2 & 33.3 & 0.464 & 0.142 & 74.7 & 2.2 \\
Low veg.    &     396 & 36.6 & 32.3 & 0.442 & 0.116 & 76.3 & 1.8 \\
Clutter     &     279 & 55.6 & 47.3 & 0.550 & 0.207 & 49.3 & 1.5 \\
\textit{Global} & \textit{1{,}392} & \textit{61.9} & \textit{54.0} & \textit{0.586} & \textit{0.175} & \textit{75.8} & --- \\
\midrule
\multicolumn{8}{c}{\footnotesize\textit{Agri-BR (Dense Grid, $T = 1{,}000$, 4{,}533 segments, 98.1\% coverage)}} \\
\midrule
Pivot irrig. &    508 & 100.0 & 100.0 & 0.996 & 0.979 & 99.7 & 1.0 \\
Crop fields & 2{,}758 & 95.2 & 94.8 & 0.945 & 0.881 & 99.6 & 1.0 \\
Lakes       &      73 & 89.0 & 89.0 & 0.888 & 0.773 & 94.0 & 1.2 \\
\textit{Global} & \textit{3{,}339} & \textit{95.8} & \textit{95.4} & \textit{0.951} & \textit{0.894} & \textit{99.5} & --- \\
\bottomrule
\end{tabular*}
\end{table}

\FloatBarrier
\subsection{Baseline Comparison}
\label{sec:res_baselines}

To test whether SAM2-based segments are better than traditional alternatives, Remote SAMsing is compared against SLIC \citep{achanta2012slic}, Felzenszwalb \citep{felzenszwalb2004efficient}, and SamGeo2 \citep{wu2023samgeo}, with SLIC and Felzenszwalb calibrated to produce a similar segment count (Table~\ref{tab:baselines}). 

\begin{figure}[htbp]
    \centering
    \includegraphics[width=0.75\textwidth]{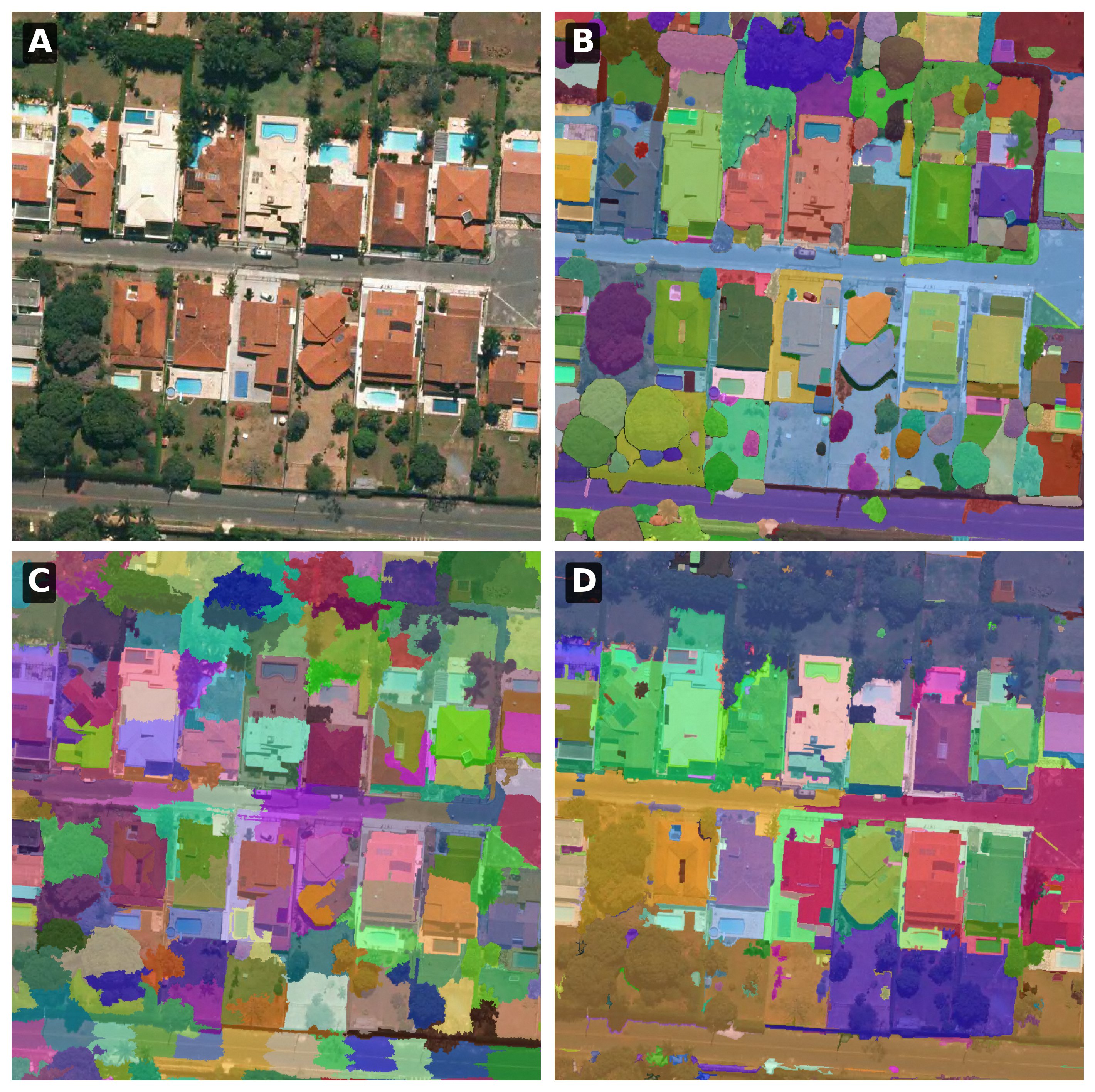}
    \caption{Visual comparison of segmentation methods on BSB-1: (A) original image, (B) Remote SAMsing ($T = 250$), (C) SLIC, and (D) Felzenszwalb.}
    \label{fig:baselines_visual}
\end{figure}

Remote SAMsing achieves the highest Det@0.5, mIoU, and BIoU on all evaluated classes across all datasets. Regarding the BIoU, Remote SAMsing reaches 0.18--0.89 globally, while SLIC, Felzenszwalb, and SamGeo2 remain below 0.21, confirming that SAM2's learned representations place boundaries along object contours rather than spectral gradients. SamGeo2 achieves only 7--29\% Det@0.5 without the multi-pass pipeline, SLIC cannot resolve small objects (2.0\% Det@0.5 for cars on BSB-1), and Felzenszwalb follows spectral similarity rather than object structure (global BIoU 0.07--0.14). Fig.~\ref{fig:baselines_visual} shows this difference.

\begin{table}[htbp]
\centering
\caption{Baseline comparison (greedy oracle protocol). Det = Det@0.5 (\%), IoU = mean greedy oracle IoU, BIoU = Boundary IoU. Best per class in bold. SLIC: compactness = 10, $n$ = 27{,}000 / 2{,}400 / 3{,}500 (BSB-1 / Potsdam-1 / Agri-BR). Felzenszwalb: $\sigma = 0.5$, min\_size = 50, scale = 1{,}569 / 5{,}739 / 5{,}061. Parameters calibrated to approximate Remote SAMsing's segment count per dataset.}
\label{tab:baselines}
\scriptsize
\begin{tabular*}{\textwidth}{@{\extracolsep{\fill}}lrrrrrrrrrrrrr}
\toprule
 & \multicolumn{3}{c}{SamGeo2} & \multicolumn{3}{c}{SLIC} & \multicolumn{3}{c}{Felzenszwalb} & \multicolumn{3}{c}{Remote SAMsing} \\
\cmidrule(lr){2-4} \cmidrule(lr){5-7} \cmidrule(lr){8-10} \cmidrule(lr){11-13}
Class & Det & IoU & BIoU & Det & IoU & BIoU & Det & IoU & BIoU & Det & IoU & BIoU \\
\midrule
\multicolumn{13}{c}{\footnotesize\textit{BSB-1}} \\
\midrule
Buildings &  4.9 & .057 & .033 & 78.1 & .640 & .168 & 75.8 & .625 & .191 & \textbf{95.1} & \textbf{.818} & \textbf{.344} \\
Trees     &  6.4 & .062 & .048 & 49.0 & .506 & .129 & 33.6 & .450 & .106 & \textbf{87.0} & \textbf{.824} & \textbf{.618} \\
Cars      & 25.3 & .216 & .179 &  2.0 & .312 & .087 & 46.6 & .508 & .310 & \textbf{81.5} & \textbf{.728} & \textbf{.555} \\
Pools     &  4.9 & .051 & .022 & 30.9 & .458 & .068 & 47.5 & .516 & .132 & \textbf{76.8} & \textbf{.700} & \textbf{.330} \\
Courts    & 13.5 & .137 & .080 & 94.2 & .733 & .189 & 71.2 & .634 & .150 & \textbf{96.2} & \textbf{.784} & \textbf{.309} \\
Decks     &  0.0 & .009 & .008 & 40.0 & .461 & .213 & 40.0 & .450 & .281 & \textbf{80.0} & \textbf{.718} & \textbf{.434} \\
Roads     &  0.0 & .004 & .008 & 67.4 & .531 & .126 & 11.6 & .326 & .110 & \textbf{79.1} & \textbf{.651} & \textbf{.241} \\
Lakes     & 12.5 & .159 & .083 & 62.5 & .510 & .180 & 50.0 & .442 & .117 & \textbf{87.5} & \textbf{.774} & \textbf{.449} \\
Permeable &  1.0 & .012 & .008 & 46.7 & .500 & .145 & 22.8 & .404 & .118 & \textbf{79.2} & \textbf{.748} & \textbf{.538} \\
\textit{Global} & \textit{6.7} & \textit{.064} & \textit{.048} & \textit{46.7} & \textit{.500} & \textit{.129} & \textit{38.4} & \textit{.471} & \textit{.139} & \textit{\textbf{85.1}} & \textit{\textbf{.790}} & \textit{\textbf{.543}} \\
\midrule
\multicolumn{13}{c}{\footnotesize\textit{Potsdam-1}} \\
\midrule
Impervious&  4.9 & .077 & .040 & 20.6 & .368 & .067 &  8.8 & .280 & .051 & \textbf{42.2} & \textbf{.464} & \textbf{.142} \\
Building  & 26.9 & .280 & .109 & 84.6 & .718 & .127 & 30.8 & .461 & .073 & \textbf{94.9} & \textbf{.861} & \textbf{.218} \\
Low veg.  &  1.5 & .040 & .021 & 30.6 & .403 & .071 &  2.8 & .271 & .036 & \textbf{36.6} & \textbf{.442} & \textbf{.116} \\
Tree      &  7.9 & .087 & .020 & 71.8 & .559 & .047 & 45.4 & .463 & .025 & \textbf{73.2} & \textbf{.582} & \textbf{.071} \\
Car       & 71.2 & .629 & .277 & 40.9 & .476 & .052 & 46.3 & .528 & .105 & \textbf{93.0} & \textbf{.818} & \textbf{.346} \\
Clutter   & 15.4 & .150 & .066 & 16.1 & .356 & .061 & 22.9 & .398 & .114 & \textbf{55.6} & \textbf{.550} & \textbf{.207} \\
\textit{Global} & \textit{20.1} & \textit{.196} & \textit{.083} & \textit{40.2} & \textit{.454} & \textit{.063} & \textit{25.4} & \textit{.394} & \textit{.065} & \textit{\textbf{61.9}} & \textit{\textbf{.586}} & \textit{\textbf{.175}} \\
\midrule
\multicolumn{13}{c}{\footnotesize\textit{Agri-BR}} \\
\midrule
Pivots    & 85.2 & .840 & .601 & 97.4 & .795 & .212 & 92.9 & .662 & .137 & \textbf{100} & \textbf{.996} & \textbf{.979} \\
Crops     & 19.1 & .189 & .135 & 53.4 & .541 & .128 & 30.9 & .451 & .104 & \textbf{95.2} & \textbf{.945} & \textbf{.881} \\
Lakes     & 23.3 & .227 & .144 & 35.6 & .464 & .076 & 64.4 & .642 & .226 & \textbf{89.0} & \textbf{.888} & \textbf{.773} \\
\textit{Global} & \textit{29.2} & \textit{.289} & \textit{.206} & \textit{59.7} & \textit{.578} & \textit{.140} & \textit{41.0} & \textit{.487} & \textit{.111} & \textit{\textbf{95.8}} & \textit{\textbf{.951}} & \textit{\textbf{.894}} \\
\bottomrule
\end{tabular*}
\end{table}

\FloatBarrier
\subsection{Scalability}
\label{sec:res_scalability}

\begin{figure}[!h]
    \centering
    \includegraphics[width=\textwidth]{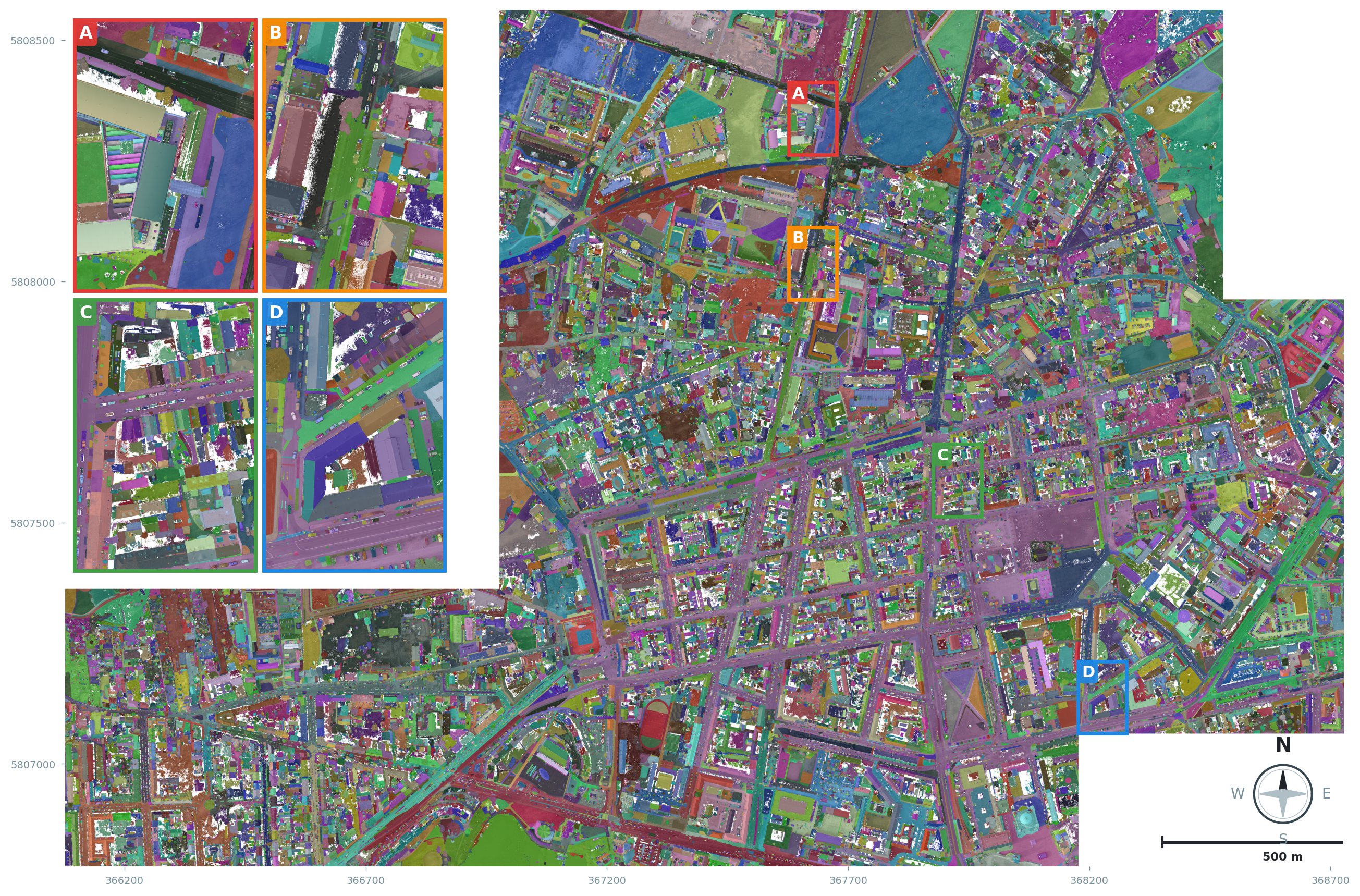}
    \caption{Full Potsdam mosaic segmentation ($36{,}000 \times 54{,}000$ pixels), with four zoomed panels (A, B, C, and D) showing segment detail in four regions of the mosaic, with matching colored rectangles indicating the source locations.}
    \label{fig:potsdam_full}
\end{figure}

Remote SAMsing processed the full Potsdam mosaic ($36{,}000 \times 54{,}000$ pixels, 1.94 billion pixels, 37 patches with ground truth) using the default configuration (Fig.~\ref{fig:potsdam_full}). The pipeline produced 124{,}180 segments with 97.0\% mean coverage, 81.8\% global ASA, and 60.7\% global Det@0.5 across 40{,}618 object instances.

\pagebreak
Per-class results confirm that quality does not degrade with image size: buildings reach 89.4\% Det@0.5 (vs.\ 94.9\% on the individual Potsdam-1 patch), cars 93.2\% (vs.\ 93.0\%), and BIoU remains high for discrete objects (buildings 0.85, cars 0.91). Amorphous classes follow the same pattern as individual patches, with low Det@0.5 (impervious 43.5\%, low vegetation 42.4\%) but high ASA (89.5\%, 80.7\%). The boundary merge handled all 1{,}944 tiles, including boundaries between adjacent Potsdam patches not present in the original dataset. Processing required approximately 20 hours on a single GPU.

\FloatBarrier
\section{Discussion}
\label{sec:discussion}

\subsection{Pipeline Components and the Coverage-Detection Trade-off}
\label{sec:disc_components}

The ablation study shows that coverage and detection quality are governed by different mechanisms. The black mask and adaptive threshold decay increase coverage but introduce large, imprecise segments in later passes, leading to a trade-off between coverage and detection accuracy (e.g., 91\% vs.\ 80\% coverage, with 56\% vs.\ 65\% Det@0.5 on BSB-1). The black mask is critical for coverage efficiency, preventing redundant mask generation over already-segmented regions. This behavior arises from progressive scene simplification: by removing accepted segments, each pass exposes previously undetectable objects. This differs from single-pass strategies, which accept noisier masks without improving coverage. SamGeo2's low coverage (6--27\%) confirms that without iterative simplification, SAM2's AMG leaves large portions of the image unsegmented. These results position Remote SAMsing as a pipeline-level solution, complementing approaches that improve per-object mask quality: a domain-adapted SAM2 could be used within the same pipeline. While conceptually related to iterative elimination strategies \citep{shepherd2019large_scale}, the proposed method generalizes across scenes without scene-specific parameterization.

\subsection{Scene-Dependent Configuration}
\label{sec:disc_config}

Tile size $T$ acts as an implicit scale parameter that resolves the coverage-detection trade-off. Since SAM2 resizes inputs to $1{,}024 \times 1{,}024$ pixels, smaller tiles magnify objects in feature space, improving detection of small objects. Reducing $T$ from $1{,}000$ to $250$ increases BSB-1 Det@0.5 from 56\% to 85\%, outperforming SAM2's native multi-scale mechanism (crop\_n\_layers), which yields only 6~pp at comparable cost. This effect is scene-dependent. On Agri-BR, where fields span thousands of pixels, $T = 1{,}000$ already yields high quality, and smaller tiles degrade performance due to merge chaining. On Potsdam-1, detection is limited by amorphous land-cover classes with diffuse boundaries, highlighting that SAM2 performs better on discrete objects than on stuff classes. Across datasets, configurations covering approximately 50~m per tile produce stable results, providing a practical guideline. These trends generalize to large-scale processing: the full Potsdam mosaic achieves 81.8\% ASA and 60.7\% Det@0.5 across 40{,}618 objects, with per-class rates (buildings 89.4\%, cars 93.2\%) consistent with individual patches.

\subsection{Comparison with Traditional Segmentation Methods}
\label{sec:disc_baselines}

The BIoU gap between Remote SAMsing and the traditional baselines reflects a fundamental difference in how boundaries are generated. SLIC and Felzenszwalb place boundaries along spectral gradients: a tree canopy may be split along internal color variation while the tree-building boundary falls within a superpixel. SAM2, trained on over one billion natural-image masks, places boundaries based on learned object structure, aligning segments with objects as a human observer would perceive them. This explains why the BIoU advantage is consistent across all three datasets despite their different spatial resolutions and spectral compositions. The comparison focuses on open-source algorithms rather than commercial tools such as eCognition \citep{baatz2000multiresolution}, which requires manual scale tuning per scene.

\subsection{Remote SAMsing as a Superpixel Generator for OBIA}
\label{sec:disc_obia}

Remote SAMsing produces over-segmentation by design: a single building may be split into roof, shadow, and facade segments. This is consistent with the superpixel paradigm in OBIA, where over-segmentation is preferred over under-segmentation because segments can be merged in downstream classification but cannot be split \citep{blaschke2010obia, benz2004ecognition, baatz2000multiresolution}. The high BIoU values indicate that over-segmentation occurs within objects rather than across object boundaries, meaning segments do not leak into adjacent classes. The gap between Det@0.5 and SS-Det@0.5 (Table~\ref{tab:perobject}) quantifies the role of the merge step: on Potsdam-1, buildings achieve 94.9\% detection with merge but only 67.9\% with single segments, confirming that boundary merging is essential for reconstructing objects that span multiple tiles. This is the fundamental requirement for OBIA: a building split into two segments is acceptable provided neither segment extends into the adjacent road.

ASA \citep{stutz2018superpixels} quantifies this directly. Remote SAMsing achieves ASA of 90.0\% on BSB-1, 75.8\% on Potsdam-1, and 99.5\% on Agri-BR, meaning that a downstream classifier assigning each segment its majority class achieves 76--99\% accuracy without additional refinement. The Agri-BR result (99.5\% ASA with MNF false-color input) stands out: SAM2 was never trained on MNF imagery, yet it segments agricultural fields with near-perfect accuracy. This suggests that SAM2's learned representations respond to spatial gradients and textural boundaries rather than spectral semantics, opening the possibility of applying the pipeline to other non-natural compositions where object boundaries are visually apparent. Supervised segmentation methods (e.g., U-Net, DeepLabV3+) can achieve higher per-class accuracy when trained on scene-specific labeled data, but require per-dataset annotation and retraining; Remote SAMsing produces competitive segmentation quality as a zero-shot, annotation-free alternative that can serve as input to such classifiers.

\subsection{Accessibility and Parameter Sensitivity}
\label{sec:disc_accessibility}

A practical barrier to adopting SAM2 for remote sensing is that the Automatic Mask Generator exposes several low-level parameters (predicted IoU threshold, stability score threshold, points-per-side grid density) whose effects on output quality are non-obvious and interact in different ways depending on the scene. Table~\ref{tab:config} illustrates this sensitivity: a single threshold change from $\tau = 0.93$ to $\tau = 0.70$ shifts coverage from 30\% to 77\% on BSB-1, while simultaneously altering boundary precision and detection rates in ways that vary across datasets. No single fixed threshold performs well across all scenes, and selecting an appropriate operating point requires understanding SAM2's internal filtering logic, knowledge that domain scientists in agriculture, urban planning, or ecology typically lack. Existing tools that wrap SAM2 for geospatial use, such as segment-geospatial \citep{wu2023samgeo}, pass these parameters through to the user, transferring the tuning burden rather than resolving it.

Remote SAMsing removes this dependency on expert parameter selection. The adaptive threshold decay traverses the full $\tau$ range within each tile, and the multi-pass mechanism accumulates segments from early (strict) passes before progressively relaxing thresholds to capture residual areas. The pipeline requires no knowledge of SAM2's internal thresholds to produce a complete segmentation. The default configuration ($\tau_\text{start} = 0.93$, $\tau_\text{end} = 0.60$, $T = 1{,}000$) produced 91--98\% coverage across all seven tested scenes without per-scene adjustment. The only parameter with a scene-dependent effect is tile size $T$, which controls the scale-detection trade-off discussed in Section~\ref{sec:disc_config} and can be set based on GSD and target object size without knowledge of SAM2 internals.

\subsection{Limitations}
\label{sec:disc_limitations}

SAM2 was trained on natural images and may produce suboptimal segments for data modalities far from its training distribution (e.g., SAR, thermal, hyperspectral) \citep{zhang2025sam2survey}. While the Agri-BR results with MNF false color suggest tolerance for non-natural compositions, performance on truly exotic data remains untested. Remote SAMsing produces instance-agnostic segments without class labels, so downstream classification is required to produce thematic maps, inherent to SAM2's class-agnostic design.

At 24~cm GSD, small objects such as cars (${\sim}12 \times 12$~pixels) and rows of parked cars present a visually homogeneous texture that SAM2 segments as a single unit, limiting per-object detection. Roads and other elongated structures that span multiple tiles remain challenging for tile-based approaches, as reflected in the gap between Det@0.5 (79\%) and SS-Det@0.5 (61\%) for roads, where elongated GT polygons require merge across multiple tiles.

As discussed in Section~\ref{sec:disc_config}, tile sizes below $T = 1{,}000$ at 5~cm GSD lead to merge chaining, limiting improvements in small-object detection at high spatial resolutions. The processing time remains high: approximately 18 hours for $T = 250$ on BSB-1 (1{,}024 tiles), dominated by SAM2 inference. Since tiles are processed independently, the pipeline is inherently parallelizable across multiple GPUs, though the current implementation processes tiles sequentially. The default configuration was empirically tuned and performs well across the tested scenes, but the threshold decay range and step size may benefit from scene-specific tuning.


\section{Conclusion}
\label{sec:conclusion}

The coverage and boundary fragmentation problems that prevent SAM2 from segmenting large remote sensing images are not limitations of the model itself but of how it is deployed. Remote SAMsing resolves both at the pipeline level, without modifying SAM2's architecture or requiring training data, taking a fundamentally different approach from work that fine-tunes or adapts SAM2 for individual patches.

The ablation study confirms that progressive scene simplification through black masking is the central mechanism behind the coverage gains, and that the quality-coverage trade-off can be navigated automatically rather than requiring manual threshold selection. The evaluation across seven scenes, three spatial resolutions, and two spectral compositions establishes tile size as the single most impactful configuration parameter, functioning as an implicit scale lever that outperforms SAM2's own multi-scale inference. The consistent per-class results on the 1.94 billion pixel Potsdam mosaic confirm that segmentation quality does not degrade with image size, establishing that the pipeline operates at production scale.

These findings have a direct practical implication: domain scientists in agriculture, urban planning, and ecology can apply SAM2 to arbitrarily large images without understanding its internal parameters. Since Remote SAMsing is model-agnostic, future foundation models can be integrated directly. Remaining challenges include improving detection of amorphous land cover classes and mitigating merge chaining at high spatial resolutions through hierarchical multi-scale processing.


\section*{CRediT authorship contribution statement}
\textbf{Osmar Luiz Ferreira de Carvalho:} Conceptualization, Methodology, Software, Validation, Formal analysis, Investigation, Data curation, Writing -- original draft, Visualization. \textbf{Osmar Ab\'{i}lio de Carvalho J\'{u}nior:} Supervision, Writing -- review \& editing, Project administration, Funding acquisition, Resources. \textbf{Anesmar Olino de Albuquerque:} Data curation, Validation, Visualization, Resources. \textbf{Daniel Guerreiro e Silva:} Conceptualization, Supervision, Writing -- review \& editing, Project administration, Funding acquisition, Resources.

\section*{Declaration of competing interest}
The authors declare that they have no known competing financial interests or personal relationships that could have appeared to influence the work reported in this paper.

\section*{Data availability}
The ISPRS Potsdam dataset is publicly available from the ISPRS website. The Bras\'{i}lia and Agri-BR datasets are available upon reasonable request. The Remote SAMsing source code is available at \url{https://github.com/osmarluiz/sam-mosaic}.

\section*{Acknowledgements}
The authors are grateful for financial support of the grant from the National Council for Scientific and Technological Development (CNPq) for Professor Osmar Ab\'{i}lio de Carvalho J\'{u}nior. This study was partly financed by the Coordination for the Improvement of Higher Education Personnel (CAPES) (Finance Code 001), Research Support Foundation of the Federal District (FAPDF) (Project 00193.00002237/2022-92), and CNPq (research projects 312608/2021-7, 421413/2023-9, and 421291/2022-2).

\bibliographystyle{elsarticle-harv}
\bibliography{references}

@inproceedings{kirillov2023segment,
  title={Segment Anything},
  author={Kirillov, Alexander and Mintun, Eric and Ravi, Nikhila and Mao, Hanzi and Rolland, Chloe and Gustafson, Laura and Xiao, Tete and Whitehead, Spencer and Berg, Alexander C and Lo, Wan-Yen and others},
  booktitle={Proc. IEEE/CVF Int. Conf. Comput. Vis. (ICCV)},
  pages={3992--4003},
  year={2023},
  doi={10.1109/ICCV51070.2023.00371}
}

@inproceedings{ravi2024sam2,
  title={SAM 2: Segment Anything in Images and Videos},
  author={Ravi, Nikhila and Gabeur, Valentin and Hu, Yuan-Ting and Hu, Ronghang and Ryali, Chaitanya and Ma, Tengyu and Khedr, Haitham and R{\"a}dle, Roman and Rolland, Chloe and Gustafson, Laura and others},
  booktitle={Int. Conf. Learn. Represent. (ICLR)},
  year={2025}
}

@article{achanta2012slic,
  title={SLIC Superpixels Compared to State-of-the-Art Superpixel Methods},
  author={Achanta, Radhakrishna and Shaji, Appu and Smith, Kevin and Lucchi, Aurelien and Fua, Pascal and S{\"u}sstrunk, Sabine},
  journal={IEEE Trans. Pattern Anal. Mach. Intell.},
  volume={34},
  number={11},
  pages={2274--2282},
  year={2012},
  doi={10.1109/TPAMI.2012.120}
}

@article{stutz2018superpixels,
  title={Superpixels: An Evaluation of the State of the Art},
  author={Stutz, David and Hermans, Alexander and Leibe, Bastian},
  journal={Comput. Vis. Image Underst.},
  volume={166},
  pages={1--27},
  year={2018},
  doi={10.1016/j.cviu.2017.03.007}
}

@article{blaschke2010obia,
  title={Object Based Image Analysis for Remote Sensing},
  author={Blaschke, Thomas},
  journal={ISPRS J. Photogramm. Remote Sens.},
  volume={65},
  number={1},
  pages={2--16},
  year={2010},
  doi={10.1016/j.isprsjprs.2009.06.004}
}

@article{felzenszwalb2004efficient,
  title={Efficient Graph-Based Image Segmentation},
  author={Felzenszwalb, Pedro F and Huttenlocher, Daniel P},
  journal={Int. J. Comput. Vis.},
  volume={59},
  number={2},
  pages={167--181},
  year={2004},
  doi={10.1023/B:VISI.0000022288.19776.77}
}

@article{chen2024rsprompter,
  title={RSPrompter: Learning to Prompt for Remote Sensing Instance Segmentation Based on Visual Foundation Model},
  author={Chen, Keyan and Liu, Chenyang and Chen, Hao and Zhang, Haotian and Li, Wenyuan and Zou, Zhengxia and Shi, Zhenwei},
  journal={IEEE Trans. Geosci. Remote Sens.},
  volume={62},
  pages={1--17},
  year={2024},
  doi={10.1109/TGRS.2024.3356074}
}

@article{osco2023sam_rs_review,
  title={The Segment Anything Model (SAM) for Remote Sensing Applications: From Zero to One Shot},
  author={Osco, Lucas Prado and Wu, Qiusheng and de Lemos, Eduardo Lopes and Gon{\c{c}}alves, Wesley Nunes and Ramos, Ana Paula Marques and Li, Jonathan and Marcato Junior, Jos{\'e}},
  journal={Int. J. Appl. Earth Obs. Geoinf.},
  volume={124},
  pages={103540},
  year={2023},
  doi={10.1016/j.jag.2023.103540}
}

@article{wang2024samrs,
  title={SAMRS: Scaling-up Remote Sensing Segmentation Dataset with Segment Anything Model},
  author={Wang, Di and Zhang, Jing and Du, Bo and Xu, Minqiang and Liu, Lin and Tao, Dacheng and Zhang, Liangpei},
  journal={Adv. Neural Inf. Process. Syst.},
  volume={36},
  pages={8815--8827},
  year={2023}
}

@inproceedings{ren2024segment_anything_everywhere,
  title={Segment Anything, From Space?},
  author={Ren, Simiao and Luzi, Francesco and Lahrichi, Saad and Kassaw, Kaleb and Collins, Leslie M and Bradbury, Kyle and Malof, Jordan M},
  booktitle={Proc. IEEE/CVF Winter Conf. Appl. Comput. Vis. (WACV)},
  pages={8340--8350},
  year={2024},
  doi={10.1109/WACV57701.2024.00817}
}

@article{wu2023samgeo,
  title={samgeo: A Python package for segmenting geospatial data with the Segment Anything Model (SAM)},
  author={Wu, Qiusheng and Osco, Lucas Prado},
  journal={J. Open Source Softw.},
  volume={8},
  number={89},
  pages={5663},
  year={2023},
  doi={10.21105/joss.05663}
}

@article{lassalle2015scalable,
  title={A Scalable Tile-Based Framework for Region-Merging Segmentation},
  author={Lassalle, Pierre and Inglada, Jordi and Michel, Julien and Grizonnet, Manuel and Malik, Julien},
  journal={IEEE Trans. Geosci. Remote Sens.},
  volume={53},
  number={10},
  pages={5473--5485},
  year={2015},
  doi={10.1109/TGRS.2015.2422848}
}

@software{zhao2023geosam,
  author={Zhao, Zhuoyi and Fan, Chengyan and Liu, Lin},
  title={Geo SAM: A QGIS Plugin Using Segment Anything Model (SAM) to Accelerate Geospatial Image Segmentation},
  year={2023},
  publisher={Zenodo},
  version={1.1.0},
  doi={10.5281/zenodo.8191039}
}

@article{ding2024samcd,
  title={Adapting Segment Anything Model for Change Detection in VHR Remote Sensing Images},
  author={Ding, Lei and Zhu, Kun and Peng, Daifeng and Tang, Hao and Yang, Kuiwu and Bruzzone, Lorenzo},
  journal={IEEE Trans. Geosci. Remote Sens.},
  volume={62},
  pages={1--11},
  year={2024},
  doi={10.1109/TGRS.2024.3368168}
}

@article{wan2025samrs_survey,
  title={A Systematic Survey and Meta-Analysis of the Segment Anything Model in Remote Sensing Image Processing: Challenges, Advances, Applications, and Opportunities},
  author={Wan, Zhipeng and Wang, Sheng and Han, Wei and Wang, Yuewei and Huang, Xiaohui and Zhang, Xiaohan and Chen, Xiaodao and Chen, Yunliang},
  journal={ISPRS J. Photogramm. Remote Sens.},
  volume={229},
  pages={436--466},
  year={2025},
  doi={10.1016/j.isprsjprs.2025.08.023}
}

@inproceedings{xiong2024efficientsam,
  title={EfficientSAM: Leveraged Masked Image Pretraining for Efficient Segment Anything},
  author={Xiong, Yunyang and Varadarajan, Bala and Wu, Lemeng and Xiang, Xiaoyu and Xiao, Fanyi and Zhu, Chenchen and Dai, Xiaoliang and Wang, Dilin and Sun, Fei and Iandola, Forrest and Krishnamoorthi, Raghuraman and Chandra, Vikas},
  booktitle={Proc. IEEE/CVF Conf. Comput. Vis. Pattern Recognit. (CVPR)},
  pages={16111--16121},
  year={2024},
  doi={10.1109/CVPR52733.2024.01525}
}

@article{zhao2023fastsam,
  title={Fast Segment Anything},
  author={Zhao, Xu and Ding, Wenchao and An, Yongqi and Du, Yinglong and Yu, Tao and Li, Min and Tang, Ming and Wang, Jinqiao},
  journal={arXiv preprint arXiv:2306.12156},
  year={2023}
}

@article{bommasani2021foundation,
  title={On the Opportunities and Risks of Foundation Models},
  author={Bommasani, Rishi and Hudson, Drew A and Adeli, Ehsan and Altman, Russ and Arber, Simran and von Arx, Sydney and Bernstein, Michael S and Bohg, Jeannette and Bosselut, Antoine and Brunskill, Emma and others},
  journal={arXiv preprint arXiv:2108.07258},
  year={2021}
}

@inproceedings{kirillov2019panoptic,
  title={Panoptic Segmentation},
  author={Kirillov, Alexander and He, Kaiming and Girshick, Ross and Rother, Carsten and Doll{\'a}r, Piotr},
  booktitle={Proc. IEEE/CVF Conf. Comput. Vis. Pattern Recognit. (CVPR)},
  pages={9396--9405},
  year={2019},
  doi={10.1109/CVPR.2019.00963}
}

@inproceedings{cheng2021boundary,
  title={Boundary {IoU}: Improving Object-Centric Image Segmentation Evaluation},
  author={Cheng, Bowen and Girshick, Ross and Doll{\'a}r, Piotr and Berg, Alexander C and Kirillov, Alexander},
  booktitle={Proc. IEEE/CVF Conf. Comput. Vis. Pattern Recognit. (CVPR)},
  pages={15329--15337},
  year={2021},
  doi={10.1109/CVPR46437.2021.01508}
}

@article{tarjan1975union,
  title={Efficiency of a Good But Not Linear Set Union Algorithm},
  author={Tarjan, Robert Endre},
  journal={J. ACM},
  volume={22},
  number={2},
  pages={215--225},
  year={1975},
  doi={10.1145/321879.321884}
}

@inproceedings{ryali2023hiera,
  title={Hiera: A Hierarchical Vision Transformer without the Bells-and-Whistles},
  author={Ryali, Chaitanya and Hu, Yuan-Ting and Bolya, Daniel and Wei, Chen and Fan, Haoqi and Huang, Po-Yao and Aggarwal, Vaibhav and Chowdhury, Arkabandhu and Poursaeed, Omid and Hoffman, Judy and Malik, Jitendra and Li, Yanghao and Feichtenhofer, Christoph},
  booktitle={Proc. Int. Conf. Mach. Learn. (ICML)},
  pages={29441--29454},
  year={2023}
}

@inproceedings{dosovitskiy2021vit,
  title={An Image is Worth 16x16 Words: Transformers for Image Recognition at Scale},
  author={Dosovitskiy, Alexey and Beyer, Lucas and Kolesnikov, Alexander and Weissenborn, Dirk and Zhai, Xiaohua and Unterthiner, Thomas and Dehghani, Mostafa and Minderer, Matthias and Heigold, Georg and Gelly, Sylvain and Uszkoreit, Jakob and Houlsby, Neil},
  booktitle={Int. Conf. Learn. Represent. (ICLR)},
  year={2021}
}

@article{green1988mnf,
  title={A Transformation for Ordering Multispectral Data in Terms of Image Quality with Implications for Noise Removal},
  author={Green, Andrew A. and Berman, Mark and Switzer, Paul and Craig, Maurice D.},
  journal={IEEE Trans. Geosci. Remote Sens.},
  volume={26},
  number={1},
  pages={65--74},
  year={1988},
  doi={10.1109/36.3001}
}

@article{xiao2025foundation_rs,
  title={Foundation Models for Remote Sensing and Earth Observation: A Survey},
  author={Xiao, Aoran and Xuan, Weihao and Wang, Junjue and Huang, Jiaxing and Tao, Dacheng and Lu, Shijian and Yokoya, Naoto},
  journal={IEEE Geosci. Remote Sens. Mag.},
  volume={13},
  pages={297--324},
  year={2025},
  doi={10.1109/MGRS.2025.3576766}
}

@article{rottensteiner2014isprs,
  title={Results of the {ISPRS} Benchmark on Urban Object Detection and {3D} Building Reconstruction},
  author={Rottensteiner, Franz and Sohn, Gunho and Gerke, Markus and Wegner, Jan Dirk and Breitkopf, Uwe and Jung, Jacek},
  journal={ISPRS J. Photogramm. Remote Sens.},
  volume={93},
  pages={256--271},
  year={2014},
  doi={10.1016/j.isprsjprs.2013.10.004}
}

@article{benz2004ecognition,
  title={Multi-resolution, Object-oriented Fuzzy Analysis of Remote Sensing Data for {GIS}-ready Information},
  author={Benz, Ursula C. and Hofmann, Peter and Willhauck, Gregor and Lingenfelder, Iris and Heynen, Markus},
  journal={ISPRS J. Photogramm. Remote Sens.},
  volume={58},
  number={3--4},
  pages={239--258},
  year={2004},
  doi={10.1016/j.isprsjprs.2003.10.002}
}

@incollection{hay2008geobia,
  title={Geographic Object-Based Image Analysis ({GEOBIA}): A New Name for a New Discipline},
  author={Hay, Geoffrey J. and Castilla, Guillermo},
  booktitle={Object-Based Image Analysis},
  editor={Blaschke, Thomas and Lang, Stefan and Hay, Geoffrey J.},
  pages={75--89},
  year={2008},
  publisher={Springer},
  doi={10.1007/978-3-540-77058-9_4}
}

@article{blaschke2014geobia,
  title={Geographic Object-Based Image Analysis -- Towards a New Paradigm},
  author={Blaschke, Thomas and Hay, Geoffrey J. and Kelly, Maggi and Lang, Stefan and Hofmann, Peter and Addink, Elisabeth and Feitosa, Raul Queiroz and van der Meer, Freek and van der Werff, Harald and van Coillie, Frieke and Tiede, Dirk},
  journal={ISPRS J. Photogramm. Remote Sens.},
  volume={87},
  pages={180--191},
  year={2014},
  doi={10.1016/j.isprsjprs.2013.09.014}
}

@inproceedings{ren2003superpixel,
  title={Learning a Classification Model for Segmentation},
  author={Ren, Xiaofeng and Malik, Jitendra},
  booktitle={Proc. IEEE Int. Conf. Comput. Vis. (ICCV)},
  pages={10--17},
  year={2003},
  doi={10.1109/ICCV.2003.1238308}
}

@incollection{baatz2000multiresolution,
  title={Multiresolution Segmentation: An Optimization Approach for High Quality Multi-scale Image Segmentation},
  author={Baatz, Martin and Sch{\"a}pe, Arno},
  booktitle={Angew. Geogr. Informationsverarb. XII},
  editor={Strobl, Josef and Blaschke, Thomas and Griesebner, Gerald},
  pages={12--23},
  year={2000},
  publisher={Wichmann, Heidelberg}
}

@article{shepherd2019large_scale,
  title={Operational Large-Scale Segmentation of Imagery Based on Iterative Elimination},
  author={Shepherd, John D. and Bunting, Peter and Dymond, John R.},
  journal={Remote Sens.},
  volume={11},
  number={6},
  pages={658},
  year={2019},
  doi={10.3390/rs11060658}
}

@article{zhu2017deep_rs,
  title={Deep Learning in Remote Sensing: A Comprehensive Review and List of Resources},
  author={Zhu, Xiao Xiang and Tuia, Devis and Mou, Lichao and Xia, Gui-Song and Zhang, Liangpei and Xu, Feng and Fraundorfer, Friedrich},
  journal={IEEE Geosci. Remote Sens. Mag.},
  volume={5},
  number={4},
  pages={8--36},
  year={2017},
  doi={10.1109/MGRS.2017.2762307}
}

@article{clinton2010segmentation_eval,
  title={Accuracy Assessment Measures for Object-based Image Segmentation Goodness},
  author={Clinton, Nicholas and Holt, Ashley and Scarborough, James and Yan, Li and Gong, Peng},
  journal={Photogramm. Eng. Remote Sens.},
  volume={76},
  number={3},
  pages={289--299},
  year={2010},
  doi={10.14358/PERS.76.3.289}
}

@article{frazier2021planet,
  title={A Technical Review of Planet Smallsat Data: Practical Considerations for Processing and Using PlanetScope Imagery},
  author={Frazier, Amy E. and Hemingway, Benjamin L.},
  journal={Remote Sens.},
  volume={13},
  number={19},
  pages={3930},
  year={2021},
  doi={10.3390/rs13193930}
}

@article{carvalho2021instance,
  title={Instance Segmentation for Large, Multi-Channel Remote Sensing Imagery Using Mask-RCNN and a Mosaicking Approach},
  author={de Carvalho, Osmar Luiz Ferreira and de Carvalho J{\'u}nior, Osmar Ab{\'\i}lio and de Albuquerque, Anesmar Olino and de Bem, Pablo Pozzobon and Silva, Cristiano Rosa and Ferreira, Pedro Henrique Guimar{\~a}es and de Moura, Rebeca dos Santos and Gomes, Roberto Arnaldo Trancoso and Guimar{\~a}es, Renato Fontes and Borges, Dibio Leandro},
  journal={Remote Sens.},
  volume={13},
  number={1},
  pages={39},
  year={2021},
  doi={10.3390/rs13010039}
}

@article{huang2018tiling,
  title={Tiling and Stitching Segmentation Output for Remote Sensing: Basic Challenges and Recommendations},
  author={Huang, Bohao and Reichman, Daniel and Collins, Leslie M and Bradbury, Kyle and Malof, Jordan M},
  journal={arXiv preprint arXiv:1805.12219},
  year={2018}
}

@article{lv2025deepmerge,
  title={Deep Merge: Deep-Learning-Based Region Merging for Remote Sensing Image Segmentation},
  author={Lv, Xianwei and Persello, Claudio and Li, Wangbin and Huang, Xiao and Ming, Dongping and Stein, Alfred},
  journal={IEEE Trans. Geosci. Remote Sens.},
  volume={63},
  pages={1--20},
  year={2025},
  doi={10.1109/TGRS.2025.3544549}
}

@article{zhang2025sam2survey,
  title={SAM2 for Image and Video Segmentation: A Comprehensive Survey},
  author={Zhang, Jiaxing and Tang, Hao},
  journal={arXiv preprint arXiv:2503.12781},
  year={2025}
}

@article{liu2026asamps,
  title={An Adaptive Segment Anything Model 2 ({SAM2}) for Planted Field Segmentation from Remote Sensing Imagery},
  author={Liu, Shuaijun and Dong, Qi and Chen, Xuehong and Dong, Xiuchun and Huang, Ping and Yang, Peng and Chen, Jin},
  journal={Int. J. Digit. Earth},
  volume={19},
  number={1},
  year={2026},
  doi={10.1080/17538947.2026.2645885}
}

@inproceedings{fan2025stablesam,
  title={Stable Segment Anything Model},
  author={Fan, Qi and Tao, Xin and Ke, Lei and Ye, Mingqiao and Zhang, Yuan and Wan, Pengfei and Wang, Zhongyuan and Tai, Yu-Wing and Tang, Chi-Keung},
  booktitle={Int. Conf. Learn. Represent. (ICLR)},
  year={2025}
}

@inproceedings{lin2025samrefiner,
  title={{SAMRefiner}: Taming Segment Anything Model for Universal Mask Refinement},
  author={Lin, Yuqi and Li, Hengjia and Shao, Wenqi and Yang, Zheng and Zhao, Jun and He, Xiaofei and Luo, Ping and Zhang, Kaipeng},
  booktitle={Int. Conf. Learn. Represent. (ICLR)},
  year={2025}
}

@article{li2024dl_seg_review,
  title={A Review of Remote Sensing Image Segmentation by Deep Learning Methods},
  author={Li, Jiangyun and Cai, Yuanxiu and Li, Qing and others},
  journal={Int. J. Digit. Earth},
  volume={17},
  number={1},
  year={2024},
  doi={10.1080/17538947.2024.2328827}
}

@article{carvalho2022panoptic,
  title={Panoptic Segmentation Meets Remote Sensing},
  author={de Carvalho, Osmar Luiz Ferreira and de Carvalho J{\'u}nior, Osmar Ab{\'\i}lio and Silva, Cristiano Rosa e and de Albuquerque, Anesmar Olino and Santana, Nickolas Castro and Borges, Dibio Leandro and Gomes, Roberto Arnaldo Trancoso and Guimar{\~a}es, Renato Fontes},
  journal={Remote Sens.},
  volume={14},
  number={4},
  pages={965},
  year={2022},
  doi={10.3390/rs14040965}
}

@inproceedings{walther2025superpixelanything,
  title={Superpixel Anything: A General Object-Based Framework for Accurate yet Regular Superpixel Segmentation},
  author={Walther, Julien and Giraud, R{\'e}mi and Cl{\'e}ment, Micha{\"e}l},
  booktitle={Br. Mach. Vis. Conf. (BMVC)},
  year={2025}
}

\end{document}